%% file: main.tex
\documentclass[sigconf]{acmart}

\usepackage{enumitem}
\usepackage{subcaption}

\copyrightyear{2023}
\acmYear{2023}
\setcopyright{rightsretained}
\acmConference[FAccT '23]{2023 ACM Conference on Fairness, Accountability, and Transparency}{June 12--15, 2023}{Chicago, IL, USA}
\acmBooktitle{2023 ACM Conference on Fairness, Accountability, and Transparency (FAccT '23), June 12--15, 2023, Chicago, IL, USA}
\acmDOI{10.1145/3593013.3594114}
\acmISBN{979-8-4007-0192-4/23/06}

\acmJournal{JACM}
\acmVolume{37}
\acmNumber{4}
\acmArticle{111}
\acmMonth{8}

\begin{document}

\title{Skin Deep: Investigating Subjectivity in Skin Tone Annotations for Computer Vision Benchmark Datasets}


\author{Teanna Barrett}
\affiliation{%
  \institution{Howard University}
  \city{Washington}
  \state{DC}
  \country{USA}}
\email{teanna.barrett@bison.howard.edu}

\author{Quan Ze Chen}
\affiliation{%
 \institution{University of Washington}
 \city{Seattle}
 \state{WA}
 \country{USA}}
\email{cqz@cs.washington.edu}

\author{Amy X. Zhang}
\affiliation{%
 \institution{University of Washington}
 \city{Seattle}
 \state{WA}
 \country{USA}}
\email{axz@cs.uw.edu}
\renewcommand{\shortauthors}{Barrett et al.}

\newcommand{\bug}{\mbox{\textcolor{red}{\rule{2mm}{2mm}}}}
\newcommand{\Bug}[1]{\bug \footnote{BUG: {#1}}}
\input{000_abstract}



\maketitle

\input{001_intro}
\input{002_related}
\input{003_lit_review}
\input{004_experiment}

\input{005_discussion}

\begin{acks}

This work was supported by an NSF REU Supplement under award \#2120497. We thank all of our experiment study participants for taking part in our study. We also thank the members of the Social Futures Lab at UW CSE, the UW DUB Summer REU program, and the Affective Biometrics Lab at Howard University for feedback on the work and piloting of our experiment. We are also grateful for preliminary resources provided by the Google Monk Scale team.

\end{acks}

\bibliographystyle{ACM-Reference-Format}
\bibliography{biblio}

\input{010_appendix}

\end{document}

%% file: 000_abstract.tex
\begin{abstract}
To investigate the well-observed racial disparities in computer vision systems that analyze images of humans, researchers have turned to skin tone as more objective annotation than race metadata for fairness performance evaluations. 
However, the current state of skin tone annotation procedures is highly varied. For instance, researchers use a range of untested scales and skin tone categories, have unclear annotation procedures, and provide inadequate analyses of uncertainty. In addition, little attention is paid to the positionality of the humans involved in the annotation process---both designers and annotators alike---and the historical and sociological context of skin tone in the United States.
Our work is the first to investigate the skin tone annotation process as a sociotechnical project. We surveyed recent skin tone annotation procedures and conducted annotation experiments to examine how subjective understandings of skin tone are embedded in skin tone annotation procedures. 
Our systematic literature review revealed the uninterrogated association between skin tone and race and the limited effort to analyze annotator uncertainty in current procedures for skin tone annotation in computer vision evaluation. 
Our experiments demonstrated that design decisions in the annotation procedure such as the order in which the skin tone scale is presented or additional context in the image (i.e., presence of a face) significantly affected the resulting inter-annotator agreement and individual uncertainty of skin tone annotations.
We call for greater reflexivity in the design, analysis, and documentation of procedures for evaluation using skin tone.
\end{abstract}

%% file: 001_intro.tex
\section{Introduction}
Computer vision (CV) technologies for analyzing and processing visual data about humans, such as facial recognition, body detection \cite{wilson2019predictive}, and skin condition classification \cite{kinyanjui2019estimating}, are being rapidly deployed in both everyday settings such as phone unlocking \cite{haider2022usability}, as well as high stakes situations such as criminal identification \cite{menon2023leveraging} or skin disease detection \cite{maduranga2022mobile}.
Implementations of CV in the wild have exposed major discrepancies in performance, particularly the misclassification and misidentification of Black people in the United States~\cite{ryan2021new}. While disparities in performance should already concern technologists, the material impacts of these inaccuracies along with the mistreatment of Black people by institutions such as police and healthcare systems compound the urgency of the problem.

One approach to address these issues is to test models on data disaggregated by race. The exemplary work in evaluating performance bias in face recognition~\cite{buolamwini2018gender} criticized this approach, pointing out that racial categories, which are highly contextualized by social attributes and vary widely in definition across societies around the world, are not a good fit for CV tasks. This is because these models only use general features of the face or perform skin detection to locate humans in an image. In other words, computer vision models only care about data that is skin deep. The social determinants that define a large part of race are not recorded in CV datasets.

To address the mismatch between deeply contextual race annotations and systems that only consider the visible aspects of race, some CV researchers have recently sought to use skin tone annotations instead. 
However, this newly adopted and relatively untested approach has also received criticism due to limitations of existing annotation scales \cite{buolamwini2018gender} and measures \cite{krishnapriya2022analysis}.
In particular, the discrete categories of some of these scales do not represent the full range of human skin tones. 
In addition, skin tone annotation procedures vary widely from project to project. There has been little attention placed on annotator uncertainty and disagreement and how this may vary according to different annotation procedure designs \cite{groh2022towards}. 
Despite researcher consensus that skin tone is at least a more objective annotation than race, skin tone still carries subjective social meaning beyond color values, which has yet to be explored.

In this work, we investigate skin tone annotation processes for computer vision benchmark datasets as a sociotechnical project that embeds subjective understandings of skin tone.
We first survey state-of-the-art usage of skin tone annotations in computer vision through a systematic literature review of CV papers utilizing skin tone annotations in the last five years (2017--2022). From a qualitative analysis of 50 papers from notable venues including IEEE Computer Vision and Pattern Recognition (CVPR) and Neural Information Processing Systems (NeurIPS), we find a wide variety of procedures and use of skin tone categories and an overall lack of documentation around annotator characteristics and procedures. We also identify areas for improvement in transparency around skin tone distribution and analysis of annotator uncertainty. Finally, we interrogate the conflation between skin tone and race found in a significant number of papers.

Following our literature review, we conducted an experiment in which we collected skin tone annotations while varying certain configurations of the annotation process such as the type of scale, ordering of scale levels, and type of image annotated. We also collected the self-reported skin tone of each human annotator to investigate annotator positionality on this task and used a range-based annotation method to isolate different measures of uncertainty.
After a deployment with 165 participants, we found that the order of the skin tone scale and the type of annotated image had a significant impact on the inter-rater and individual uncertainty of annotators, while we did not find a significant impact regarding scale type or annotators' self-reported skin tone. We document the results of our literature review, experiment data, and experiment code in an online repository.\footnote{\url{https://github.com/Social-Futures-Lab/skin-deep}}

Our work demonstrates the importance of reflexivity when designing and implementing skin tone annotation procedures. If the humans involved in the skin tone annotation process do not acknowledge their subjective social biases, it is difficult to ensure the reliability of skin tone annotations across the field. In turn, unclear or limited reliability of skin tone annotations can affect the robustness of fairness performance evaluations \cite{costanza2022audits} in CV and hinder the development of solutions for racialized disparities in CV models. 
We conclude with a discussion of design implications for CV researchers engaging in skin tone annotation.

%% file: 002_related.tex
\section{Background and Related Work}
\subsection{Impacts of Computer Vision for Analyzing Images of Humans}
Computer vision (CV) is a sub-domain of machine learning (ML) in which images, videos, and other visual data are analyzed to derive specific information about the inputs. While a variety of data and insights have been the focus of CV, a large portion of the field is concerned with processing humans in visual data. Efforts in human-concerned CV tasks typically aim to surveil \cite{almeida2022ethics}, identify (face recognition \cite{yucer2022measuring}), and assess physical attributes (skin condition classification \cite{kinyanjui2019estimating}). As the accuracy of these systems improve, human-concerned CV technology has increasingly been deployed in public life, including in major institutions such as education, government, and healthcare. 

While these technologies may perform well in experimental and testing environments, the implementation of CV in the wild has exposed major performance gaps. In particular, there is a clear pattern of face recognition misidentifications of Black users and subjects in the United States \cite{ryan2021new,klare2012face}. 
Black people have been unjustly arrested as a result of being misidentified by police departments' facial recognition systems \cite{ryan2021new}. Another example is the low detection rates of skin diseases on darker skin tones \cite{daneshjou2022disparities,feathers2021google}. In addition, during the height of the pandemic, remote students took exams proctored by facial analysis technology. Students with darker complexions had to shine additional light on their faces because these proctoring platforms were not detecting them \cite{swauger2020our}. In all of these examples, the disparate low performance of the CV systems mirrors and contributes to the systemic racism in every arena of American society, including policing \cite{vitale2021end}, medicine \cite{hoberman2012black}, and education \cite{swauger2020our}. As the harmful outcomes of CV technology towards Black people become more frequent and well-known, CV practitioners and researchers have a responsibility to investigate and address the racialized performance gaps in their systems \cite{birhane2021algorithmic}. 
 
 \subsection{Approaches to Investigate Disparities in Computer Vision Performance}
 The current approach in understanding and addressing performance gaps in machine learning is to test models on data disaggregated by race \cite{khan2021one}. Although most technologies do not use the protected class of race as a feature \cite{o2017weapons}, data points can have additional race labels to directly compare the performance of a model between different races. While this approach has successfully highlighted quantifiable disparities in tasks such as loan risk assessments \cite{o2017weapons}, race annotations are not ideal for evaluating disparate performance in CV tasks. The hallmark work Gender Shades \cite{buolamwini2018gender} was one of the first to evaluate performance bias in face recognition and clearly explain the unreliability of race annotations as a tool for evaluating disparity in CV. Race is more than phenotypical features and its categories are highly contextual \cite{buolamwini2018gender}. Yet, human-concerned CV operates only on surface level information. For example, many face recognition models use general features of the face to execute their tasks. Furthermore, many models utilize skin detection to locate a human in images. In contrast, when it comes to race, every society defines the boundaries of the phenotypical attributes of race differently. If these annotations are not self-reported, annotators or researchers must use their personal view of racial categories to decide how image subjects are categorized \cite{khan2021one}, which threatens the reliability of the annotations. 
 
 This creates a tension in which race annotations which, as described, are a highly contextual record of perceptions and biases, are used to evaluate systems that only contend with the surface level aspects of race. This mismatch introduces cases in which a dataset with image subjects of all the same race may have widely different appearances, or a multi-racial image dataset of subjects may have similar appearances. Limited insights can be gained from a performance evaluation disaggregated by race in these two dataset instances. If they were instead annotated by skin tone, the performance of the CV systems can be evaluated based on how these models actually ``see'' race. 

\subsection{Origin, Limitations, and Sociotechnical Aspects of Skin Tone Annotations}
The most popular measure for skin tone is the Fitzpatrick Skin Type (FST) which was created and proposed by Thomas Fitzpatrick \cite{fitzpatrick1988validity} for analyzing sun burns. This set of six color swatches was meant to only be accompanied with written descriptions of the quality of the skin and the corresponding reactions to sun exposure.  
Following the FST scale, an automatically calculated continuous measure called Individual Typology Angle (ITA) was developed to measure skin tone in relation to sun exposure sensitivity more precisely \cite{del2006relationship}. 
Even within the original scope of these skin tone measures, racial perspectives were embedded in the process. The first iteration of FST \cite{fitzpatrick1988validity} and ITA \cite{chardon1991skin} were explicitly meant for Caucasians. Even as these measures were updated, they still did not represent skin tones of Black or Asian populations as well \cite{ware2020racial}. The legacy of neglect for darker skin tones is evident in the continued lack of darker skin tones represented in dermatology image datasets \cite{lester2020absence,arosarena2015options}. Gender Shades \cite{buolamwini2018gender} was one of the first works to use skin tone annotations in CV to evaluate the performance of commercial facial recognition technologies. Since then, use of skin tone annotations in CV has grown, and new scales and measures exist such as the Monk Skin Tone (MST) scale, released by Google Research in 2022 \cite{monkscale}. MST has ten color swatches and was designed in collaboration with the sociologist Ellis Monk \cite{monk2019color} to explicitly represent a broader range of people. However, to date, there has not been a public evaluation of the MST scale or study of its implementation. 

In addition to improving the representation of skin tone scales, an understanding of the deeper sociological context of skin tone stratification is an important part of addressing the limitations of skin tone annotation. One of the primary contributors to the Monk Skin Tone scale has extensively studied how attitudes towards individuals with lighter and darker skin tones from the time of slavery in the United States persists in contemporary American society \cite{monk2021unceasing,monk2021skin,monk2019color}. Racism is integral to skin tone stratification, but skin tone also has unique and pervasive social and political impacts of its own. There is a plethora of literature on the need to transparently articulate the assumptions, biases, and perspectives that are embedded in building datasets for machine learning as a whole \cite{denton2021whose,jo2020lessons,denton2020bringing}. Given the complex social reality of skin tone in the U.S., there is a specific need to investigate the implicit and explicit assumptions, biases, and perspectives in the skin tone annotation process as well.

\subsection{Issues with Skin Tone Annotations for Computer Vision Evaluation}
While skin tone is a viable annotation attribute for computer vision, its potential is limited by the current state of skin tone annotation processes. The primary problem identified by researchers is the imprecise nature of the annotation scales and measurements \cite{buolamwini2018gender,krishnapriya2022analysis}. The discrete categories in scales like FST do not sufficiently represent the possible skin tones of human subjects. Another aspect that has not been deeply explored is annotator and annotation uncertainty. There has been some documentation and analysis of the uncertainty of skin tone annotations 
\cite{groh2021evaluating,groh2022towards,bahmani2021sreds,wilson2019predictive,celis2020implicit,toyoda2021predicting,chang2018robust}, though our literature review finds this is rare. These works primarily provide analyses such as inter-rater reliability or some other measure of consensus or (dis)agreement.
However, there is little investigation of the implicit biases or social environments that could also impact the certainty and consistency of the annotations. As many researchers have accepted skin tone as a more objective annotation than race, there has been little investigation into how skin tone annotations may also carry social meaning. Through our experiment, we provide empirical data on how different annotation processes such as differing scales and social aspects such as annotator positionality may impact annotation uncertainty.

%% file: 003_lit_review.tex
\section{Literature Review of Skin Tone Annotation in Computer Vision}
We conducted a literature review to understand the current state of skin tone annotation procedures for computer vision datasets. As part of this literature review, we identified the common measurements and annotation procedures, annotator information, race metadata collected about the dataset subjects, and uncertainty analyses for skin tone annotations in the past five years (2017--2022). 
The complete annotated literature review is available in our project repository online.

\subsection{Method}
\subsubsection{Researcher Positionality}
In line with the feminist practice of reflexivity \cite{attia2017ing}, we articulate how the identity, perspectives and social standing of the researcher informs our survey. In particular, the selection, pruning, and analysis of the skin tone annotation procedures are guided by the first author's understanding of skin tone and race. The literature review was conducted by the first author, who is Black and based in the United States. Given the positioning of the researcher, there is a deep understanding of the skin tone at the intersection of a Black, English-speaking, and Western culture. As a member of the computer science academic community, the first author is also able to approach the survey from a technical understanding of dataset collection and annotation.

\begin{table}[]
\small
\begin{center}
\begin{tabular}{ p{0.5\linewidth}|p{0.2\linewidth} | p{0.2\linewidth} }
  \toprule
  \textbf{Publication Venue} & \textbf{No. of Search Results} & \textbf{No. of Final Papers} \\ 
  \midrule
  IEEE Xplore (IWBF, BBIS, WACV, CVPR, ICB, FG, Access, +20 more) & 99 & 37\\ 
  \hline
  ACM Digital Library (FAccT, CSCW) & 38 & 4 \\ 
  \hline
  NeurIPS & 11 & 4 \\ 
  \hline
  Patterns & 4 & 1 \\ 
  \hline
  IS\&T EI & 3 & 2 \\ 
  \hline
  ECCV & 1 & 1 \\
  \hline
  Preprint (no search conducted) &   & 1 \\
  \midrule
 \textbf{ Total} & 157 & 50 \\ 
  \bottomrule
\end{tabular}
\caption{Number of papers returned from a keyword search from each publication venue, along with the resulting papers in the final set of 50.}
\label{paper_count}
\end{center}
\end{table}

\subsubsection{Identifying Skin Tone Annotation Procedures in Computer Vision Papers}
To begin, we formulated a collection of keywords related to skin tone or skin color annotation. We collected 26 initial papers via manual searching on Google Scholar for papers that included the phrase "skin tone annotation." Papers that detailed a skin tone annotation procedure in their methodology were recorded, and additional papers that also detailed skin tone annotation procedures were found through citation trails. From perusing these papers, we chose the phrases: ``skin tone'', ``skin type'', ``Fitzpatrick skin type'', ``darker skin'', and ``lighter skin''. As many top computer vision publication venues are part of IEEE, we chose to conduct a search on all of IEEE's Xplore digital library.
In addition to the IEEE venues, we included the publication venues of the original 26 papers, excluding one which was a preprint.
With the selection of keywords and conferences, the author completed an inclusive ``OR'' search of the keywords, title, and abstract in the IEEE Xplore library and the ACM Digital Library (ACM DL). We also conducted Google Scholar searches with the publication's name for venues not available through IEEE Xplore or ACM DL. 
Table~\ref{paper_count} has the full breakdown of venues and number of search results from each.

\subsubsection{Pruning}

With a corpus of over 100 papers, the first author read over the abstract and dataset sections of each paper to remove works that did not conduct skin tone annotation procedures. Of the papers that were removed, many conducted imaging or sensing experiments rather than computer vision tasks. One paper written in Portuguese was omitted due to our own language limitations.
Several makeup and fashion recommendation papers were removed because the skin tone annotations referred to the texture or condition of the skin (e.g., oily or containing acne) as opposed to complexion. 
Finally, we found many papers that did not annotate skin tone for their training or testing datasets. 
In these papers, CV tasks were paired with calls for skin tone diversity and the acknowledgement of disparate results between lighter-skin subjects and darker-skin subjects, yet ultimately the papers did not incorporate any skin tone annotations or evaluations. The prevalence of acknowledgement followed by inaction was noted as a concerning trend that disqualified a majority of the papers that were collected by the keyword and abstract search. 
After these stages of pruning, 50 papers remained that were analyzed for their skin tone annotation procedures and analysis.

\subsubsection{Qualitative Analysis}
In \autoref{dimensions-analysis}, we list all the dimensions along which we annotated the papers, including their definition and an example of what was coded.
Our dimensions were partly informed by a prior literature review done by Scheuerman et al.~\cite{scheuerman2020we} on race and gender annotations in image datasets. We adapted their approach for the topic of skin tone annotations.
First, we collected basic information about the skin tone annotation dataset, such as the CV task and the number of subjects in the dataset.
Next, we collected information about the annotation procedure, including who or what annotated the dataset and what scale was used for the range of skin tones.
We then marked whether the paper described any analysis of the skin tone annotations for uncertainty. This included any procedure or metric to measure or evaluate the (dis)agreement, consensus, or reliability of annotations.
We finally marked whether there were any race annotations or metadata intended to represent the race, ethnicity, or nationality of an image subject. 

\subsection{Results}

\def\doitems{\def\item{\par
   \noindent\hbox to1.5em{\hss$\bullet$\hss}\hangindent=1.5em }}

\begin{table*}
    \small
    \begin{center}
    \begin{tabular}{p{0.15\textwidth}|p{0.58\textwidth}|p{0.18\textwidth}}
         \toprule
         \textbf{Category}  & \textbf{Findings} & \textbf{Papers} \\
         \midrule
         Robust Skin Tone Annotation Process & 
          \doitems 
             \item  Smaller datasets, emphasis on skin tone diversity
             \item Mostly manual, use of multiple human annotators per item
             \item Use of FST scale albeit with simplifications to ``lighter'', ``darker'' categories 
             \item Reported uncertainty analyses  & \cite{groh2021evaluating,groh2022towards,bahmani2021sreds,wilson2019predictive,celis2020implicit,toyoda2021predicting,chang2018robust,krishnapriya2022analysis,howard2021reliability} \\
         \hline
         Ambiguous Skin Tone Annotation Process & 
         \doitems 
             \item  Face recognition or health classification tasks were common
             \item 1/3 were automatically annotated using ITA; the rest used human annotators, oftentimes the authors themselves
             \item Mentioned FST scale but often used custom categories instead
             \item Typically lacked information about annotation process or distributions
             \item Lacked uncertainty analyses &   \cite{hazirbas2021towards,maze2018iarpa,merler2019diversity,beal2022billion,kinyanjui2019estimating,perera2021virtual,aledhari2021multimodal,al2018remote,garcia2018pornographic,tariq2018designing,kim2022out,koshy2021complexion,mcduff2021synthetic,doshi2022fhp,spetlik2018non,al2017simultaneous,yucer2022measuring,zhang2020open,anderson2019understanding,menezes2021bias,shah2022deep,laranjeira2022seeing,mehrotra2021mitigating,park2018automatic,le2020anonfaces,liu2021metaphys,bagdasaryan2019differential,guccluturk2017reconstructing,park2021reliable,buolamwini2018gender,sixta2020fairface,cook2019demographic,whitelam2017iarpa,zheng2022automatic,molina2020reduction,muthukumar2019color,mishra2021dual,wang2016algorithmic,falcon2022image,zhang2019analysis,mcduff2019characterizing} \\
         \hline
         Skin Tone and Race Annotation Process & 
            \doitems 
             \item  Race is used to contextualize skin tone or equate with skin tone
             \item Lacked  uncertainty analyses & \cite{buolamwini2018gender,krishnapriya2022analysis,sixta2020fairface,howard2021reliability,cook2019demographic,whitelam2017iarpa,zheng2022automatic,molina2020reduction,muthukumar2019color,mishra2021dual,wang2016algorithmic,falcon2022image,zhang2019analysis,mcduff2019characterizing} \\
         \bottomrule
    \end{tabular}
    \caption{Summary of findings from analyzing prior skin tone annotation procedures and datasets.}
    \label{results}
    \end{center}
\end{table*}


We broadly grouped the papers in our literature review into three categories for further comparison, as some papers reported on or made use of certain dimensions while others did not. The first category, `Robust Skin Tone Annotation Process', refers to papers that provide a detailed account of their skin tone annotation process, annotation uncertainty analysis, and distribution of skin tones represented in their dataset.
Our second category, `Ambiguous Skin Tone Annotation Process', comprises processes that did not include one or more of the components of a robust skin tone annotation process. 
These two categories are mutually exclusive and together comprise the full set of 50 papers. Our third orthogonal category of `Skin Tone and Race Annotation Process' selects papers with processes which collected or annotated both skin tone and race data. 
For each category, we report on the dimensions and highlight trends. 
A summary of findings is in Table~\ref{results}.

\subsubsection{Robust Skin Tone Annotation Process} There were only 9 papers out of the 50 total in our literature review with a detailed description of process, uncertainty analysis, and reporting of skin tone distribution.

\textbf{Dataset}: The papers in this category were mostly conducting skin tone classification or skin condition classification and created skin tone annotations to evaluate their performance ($n=4$). 
Even for the tasks that were not explicitly about skin classification, the goal of creating a ``diverse'' training and testing dataset was a central motivation for collecting skin tone annotations. 
This prioritization of diversity in the dataset limited dataset size---for instance, Wilson et al. \cite{wilson2019predictive} noted their lack of image subjects with darker-skin tones limited the size of the dataset as they desired a balanced sample. 
Another important constraint was time and labor for the skin tone annotations. 
The largest datasets in this category of works were achieved partially due to fully automated annotations \cite{bahmani2021sreds} or completed by a large crowd of annotators \cite{celis2020implicit}. 
Excluding the two largest datasets, which have more than 200,000 images and were not collected by the authors themselves, the datasets in this category had an average size of 7,789 images.

\textbf{Procedure}: All of the works except one had human annotators involved in the process. Some combined approaches such as automated procedures to compare the work done by subject matter expert annotators and non-expert crowd work annotators~\cite{groh2022towards}. The majority of the works also had multiple annotators involved in the annotation process, and authors recognized and reported how annotations from different annotators had differences that had to be reconciled \cite{groh2022towards}.
Authors used many methods to coordinate multiple annotators. For procedures that included crowd workers, platforms such as Amazon Web Services Mechanical Turk (AWS MTurk), Scale AI, and Centaur Labs were used. Sometimes authors weighed the annotations based on annotators' experience and reliability. 
All of the manual procedures referenced the Fitzpatrick Skin Type (FST) scale. In two works, authors calculated Individual Topology Angle (ITA) values and then binned the skin tone values into six, FST-inspired categories from lighter skin tones to darker tones \cite{groh2022towards,chang2018robust}. In one work, even when human annotators provided FST annotations, the annotations were combined into two bins: ``lighter'' or ``darker'' \cite{celis2020implicit}. Finally, three papers referenced FST but had annotators place images into lighter and darker categories. 

\textbf{Uncertainty Analysis}: 
The uncertainty analysis in these works mostly operated as a tool for determining the final value, for instance, determining an annotation was reliable if it had majority consensus \cite{wilson2019predictive}. 
Another paper relied on an opaque ``dynamic consensus'' functionality of the chosen annotation platform to determine final ratings \cite{groh2021evaluating}. 
The newest iteration \cite{groh2022towards} included inter-rater reliability metrics such as Pearson correlation coefficients and qualitative confusion matrices between different annotators. 
Papers used statistical measures such as Krippendorff’s alpha \cite{toyoda2021predicting}, Cohen’s k \cite{celis2020implicit}, and normalized standard deviation \cite{bahmani2021sreds} to evaluate the overall (dis)agreement in the skin tone annotations.
While the computed metrics varied, all serve as a step towards transparency and discussion of the limitations of skin tone annotations as an exact measurement. Evaluating (dis)agreement between annotators was the common approach for understanding annotation uncertainty.

\textbf{Summary}: The papers with more robust processes valued having multiple human annotators to provide more reliable annotations. They also explained their procedures for determining consensus and many went a step further to acknowledge and evaluate (dis)agreement across  annotations. However, this annotation process was difficult to scale because of time, labor, and compensation constraints. Most critically, the use of FST as the annotation scale led to greater uncertainty and imbalance in distribution, leading authors to simplify the skin tone categories. 

\subsubsection{Ambiguous Skin Tone Annotation Process}
The majority of the papers ($n=41$) detailed their skin tone annotation process ambiguously. Many of these works did not \textbf{A}) conduct uncertainty analysis ($n=34$), \textbf{B}) include a description of the skin tone annotation process ($n=10$), \textbf{C}) mention the skin tone annotation scale that was used ($n=8$), or \textbf{d}) clearly mention who were the skin tone annotators ($n=7$). 

\textbf{Dataset}: The most common tasks were face recognition ($n=7$), health tasks such as heart rate tracking and skin condition classification ($n=8$), synthetic image generation ($n=4$), and fashion recommendations ($n=3$). 
We note that face recognition researchers have taken interest in evaluation using skin tone, after the publication of major disparities in performance~\cite{buolamwini2018gender}. 
The ongoing interest in health ML paired with dermatological roots of skin tone annotation may also explain the prevalence of health-related tasks.  
Datasets varied in size from millions of images to 5 videos. The larger to mid-sized datasets utilized previously made datasets  ($n=22$), with  CelebA \cite{liu2015deep} being the most frequently used dataset ($n=5$). 
Newly made datasets were primarily composed of online images ($n=4$) or collected in-person ($n=6$). 

\textbf{Procedure}: Some works automated their skin tone annotation processes ($n=10$), with ITA being the most popular measure. All except one paper simplified the continuous values from ITA into ranges, with at most 8 bins. 
However, similar to the robust annotation papers, the majority of the works relied on human annotators. Many manual annotations were done by the creators of the dataset ($n=7$). However, only two of these papers \cite{mcduff2021synthetic,hazirbas2021towards} described how the authors annotated skin tone. Even then, there was little to no description of any training, the positionality of the annotators, or the annotation environment  (e.g., the use of reference images or color swatches). 
The majority of these works referenced the FST scale. However, unlike the robust annotation papers, the author annotators developed custom skin tone categories based on common skin color descriptions (e.g., dark, brown, medium, fair, white). 
In a few cases, authors outsourced annotations to subject-matter-experts \cite{anderson2019understanding} or crowdworkers \cite{menezes2021bias,sixta2020fairface,whitelam2017iarpa}. 
Finally, there were a few papers that did not record information about their skin tone annotators. 
In addition, the majority of the works did not provide any information about the distribution of the skin tone categories represented in their dataset. In one paper, the authors made use of FST, finding the third FST tone was the only skin tone in their dataset \cite{spetlik2018non}; this bias was then left as is. 

\textbf{Uncertainty Analysis}: Unfortunately, none of these works reported any metrics or commentary about the uncertainty, inter-rater reliability, or disagreement of the skin tone annotations.

\textbf{Summary}: 
It is a positive sign to see papers in recent years using skin tone annotations for evaluation and sometimes noting biased distributions in their datasets.
However, the lack of information about annotation processes in the majority of the papers in our literature review demonstrates that skin tone annotations are not yet seen as an integral part of dataset quality or an important attribute to evaluate performance of CV tasks. 
This affirms the need for developing standards for conducting and reporting on skin tone annotations.

\subsubsection{Skin Tone and Race Annotation Process}
We chose to more deeply examine a subset of the papers ($n=13$) 
that we noticed blur the boundaries between more objective skin tone annotations and socially-specific race metadata.
In some cases, the presence of race metadata is not detrimental to the intended goal of evaluating skin tone representation in datasets. 
On the other hand, certain conflations of skin tone and race reinforce harmful practices of essentializing race. 

\textbf{Dataset}: All of these works used relatively small datasets. There was a fairly even distribution of computer vision tasks such as face recognition \cite{buolamwini2018gender,maze2018iarpa,mishra2021dual}, heart rate extraction \cite{wang2016algorithmic,zhang2019analysis}, and gender classification 
 \cite{molina2020reduction,muthukumar2019color}. 

\textbf{Procedure and Race Annotations}: We found three ways  skin tone and race data were used in the annotation process. The race, ethnicity, or nationality of the image subject was used to: \textbf{A}) provide general demographic information and was not involved in the skin tone annotation process, \textbf{B}) was considered in the selection of data before the skin tone annotation process, or \textbf{C}) was the skin tone scale. 
\emph{\textbf{A}) Race Metadata for Context}: The majority of works provided aggregated racial data to provide more context about the individuals in their dataset ($n=6$). 
\emph{\textbf{B}) Race Metadata for Data Selection}: Three papers used race metadata to determine which images were selected for skin tone annotation \cite{buolamwini2018gender,wang2016algorithmic} or represented in the final dataset \cite{howard2021reliability}. 
Both works provided explanation as to why the race of the image subjects were factored into the process. For example, Howard et al.~\cite{howard2021reliability} only used images of Black and White subjects because other races present in the subject pool were not well represented.
In the other paper, the authors collected images from African and Nordic countries because ``African countries typically have
darker-skinned individuals whereas Nordic countries tend to have lighter-skinned citizens'' \cite{buolamwini2018gender}. But the authors also noted in their paper the limitation of associating skin tone with race or ethnicity.

\emph{\textbf{C}) Race = Skin Tone}: Finally, four papers used race language for the skin tone annotation values \cite{falcon2022image, zhang2019analysis, mcduff2019characterizing, molina2020reduction}. In one paper, one of the skin tone annotation categories was defined as ``Asian-skin'' \cite{falcon2022image}. 
In this case, the racialized framing of the skin tone annotations contradict the more objective purpose of skin tone annotations. Although the authors refer to skin color of the modeled subjects, additional elements such as hairstyle, eye color and facial features are important to the design of their study. By only recording skin tone, the other features encoded into the subjects are obfuscated, and the complex racial content of their subjects are uninterrogated. 
In a second paper, the skin tone annotation categories were  ``Caucasians'', ``Yellows'', and ``Blacks'' \cite{zhang2019analysis}, where one skin tone had a more overtly stereotypical description that can be assumed to be an association with individuals from Asian-descent. 
This assumption is confirmed by the fact that the authors referred to the image subjects as ``White people, black people, and yellow people.'' 
Finally, another paper describes their skin tone annotation categories as ``Black'', ``South Asian'', ``Northern Asian'', and ``White'' \cite{mcduff2019characterizing}. While these  categories offer more distinction than the previous works, the racializing of the annotation still contradicts the intended objective purpose of skin tone annotations.
The authors rationalize their choice by asserting that the ethnic homogeneity of the different regions can extend to skin tone homogeneity; however, skin tone annotation is fundamentally visual.
\textbf{Uncertainty Analysis}: Only two works reported any uncertainty evaluations \cite{howard2021reliability,krishnapriya2022analysis}. These works were classified as robust annotation procedures and also notably did not use racial language in the skin tone annotation labels. As for the other works, the lack of uncertainty analysis is consistent with the larger lack of engagement with ambiguity in the annotation process. 

\textbf{Summary}: This category of papers is the clearest demonstration of why skin tone annotations can be considered a sociotechnical project. The creators' perspectives on skin tone and race were embedded throughout the whole process, both explicitly and implicitly. 

%% file: 004_experiment.tex
\section{Experimenting with Skin Tone Annotation Design}

As can be seen, works varied widely in their skin tone annotation processes when reported, though there was also a lack of reporting about these processes in many papers. We note that many of the manually deployed practices have not been tested or compared against each other. In addition, it is unclear how different design decisions may impact annotator uncertainty or (dis)agreement.

Thus, we performed an annotation experiment to investigate how different designs decisions in the annotation process may affect the resulting annotations.

\subsection{Method}
\label{sec:method}

\begin{figure*}
    \centering
    \includegraphics[width=\textwidth]{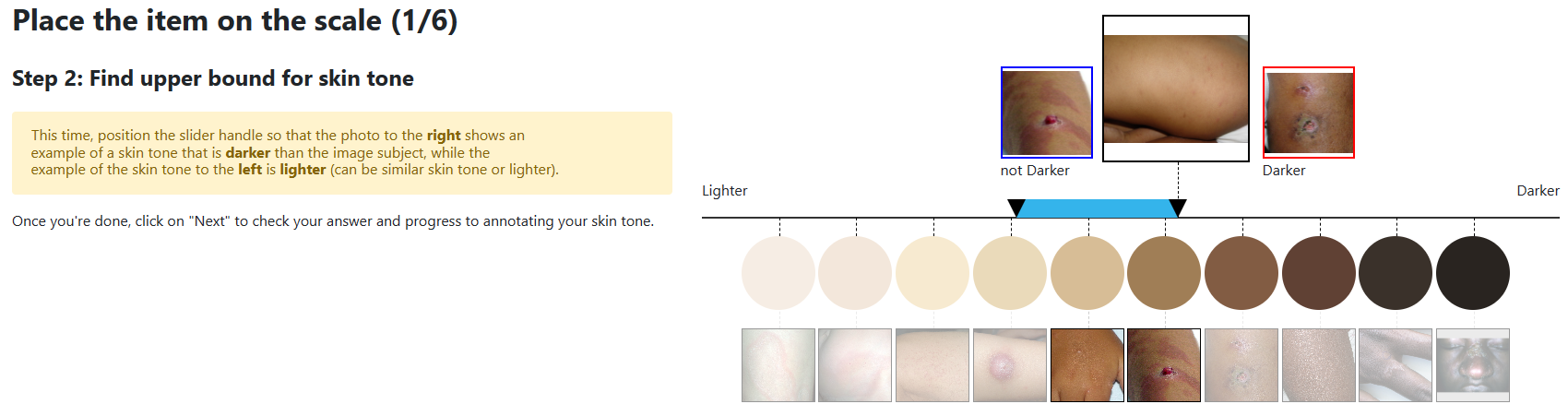}
    \caption{A screenshot of the annotation interface being used to annotate an example image in the \textsc{skin} dataset.} 
    \label{fig:annotation-tool-ui-example}
\end{figure*}

\textbf{Annotation Tool}: We based the design of our annotation tool on a range-based annotation system, Goldilocks~\cite{chen2021goldilocks}, which utilizes a two-stage process to collect range ratings on a continuous scale (Figure~\ref{fig:annotation-tool-ui-example}). One goal in our experiment was to explore limitations of the level of granularity afforded by existing scales. Thus, we used continuous ratings to give annotators full freedom in assigning their ratings while referencing a scale rather than being tied to the established levels on a specific skin tone scale. Additionally, like in Goldilocks, each annotator provided both upper and lower bounds rather than a single rating value. This allowed for a per-annotator estimate of uncertainty around each annotation, improving our insights into uncertainty beyond disagreement metrics. 
To ground the pre-existing scales (either MST~\cite{monkscale} or FST~\cite{10.1001/archderm.1988.01670060015008}), we displayed the color swatches that define the scale levels as anchors under our continuous scale. As MST and FST do not specify distances between levels, we placed anchors uniformly across our scale. We also provided an anchor in the form of an example image drawn from the dataset of the task images.

\textbf{Experiment Setup}: For an annotator on a rating task, the most important factors they will engage with are the scale and the items being rated. Thus, we set up the following conditions (denoted in \textsc{small-caps}) to test how differences here might affect the skin tone annotations produced:
\begin{itemize}[noitemsep,topsep=0pt,leftmargin=*]
    \item Scale Type: We tested the 6-point FST~\cite{10.1001/archderm.1988.01670060015008} scale (\textsc{fitz}) and the 10-point MST~\cite{monk2019color} scale (\textsc{monk}).
    \item Scale Order: Since each scale is mapped to a [0, 1] range, we tested varying the scale order to be either \textit{lighter to darker} (\textsc{ld}) where lower values represent lighter skin tones, or \textit{darker to lighter} (\textsc{dl}) where lower values represent darker skin tones. Scale values are presented left-to-right for increasing values.
    \item Image Type: We tested 2 different types of images, based on whether a whole face was visible. Images from the \textsc{skin} condition contained images of skin conditions without faces or with partial facial features~\cite{groh2021evaluating} while images from the \textsc{face} condition contained portrait photos with the subject's face~\cite{maze2018iarpa}.
\end{itemize}

In our experiments, we tested all combinations above, resulting in 8 total combinations for how the annotation tool could be configured. Across all conditions, annotators were asked to first familiarize themselves with the tool by completing a tutorial using the tool under that configuration, with feedback mechanisms on their answer similar to gated instructions used in crowd task training~\cite{liu2016effective}, before proceeding to the annotation task. 
To investigate annotator positionality, we asked annotators to mark their own skin tone along the scale using the annotation tool after completing the tutorial. Based on prior recommendations \cite{howard2021reliability}, we provide tips for assessing their skin tone.

\textbf{Image Dataset}:
We drew images from two datasets: the Fitzpatrick17k \cite{groh2022towards} (\textsc{skin}) and the IARPA Janus Benchmark-C (IJB-C) \cite{maze2018iarpa} (\textsc{face}). 
The Fitzpatrick17k dataset is a 17,000 image dataset of medical images of skin conditions from a variety of publicly available medical textbooks. The IJB-C dataset holds over 138,000 publicly available images from the Internet. Both datasets were manually annotated using FST and included annotations from crowdworkers annotators. 
We curated images from each dataset to achieve a spread across skin tones, resulting in 12 selected images for the \textsc{face} dataset and 11 images for the \textsc{skin} dataset.
Annotators were then randomly assigned to annotation task sessions where they annotated one group of 6 images under one of our 8 conditions.

\textbf{Annotator Recruitment}: 
For the main component of our annotation study, we recruited 160 U.S.-based annotators from AWS MTurk with the criteria of having completed at least 1000 tasks with a 95\% or higher approval rate.
Crowd annotators were paid \$8.50 per task (\$2.50 base pay and \$1.00 bonus per annotation) for a 30 minute task, and were not allowed to participate in more than one condition.
We conducted manual quality control checks for spam behavior and redeployed instances where this was observed. 
After removing incomplete tasks, we had a final set from 153 crowd annotators.
Annotators were asked to provide general demographic information to determine the representation of our participant sample along race and gender identities in the U.S. context. This enabled us to diagnose and correct for demographic skew in our sample.
After the AWS MTurk deployment, we noticed that our recruited annotator population skewed heavily towards those who self-reported as White (84.2\%). 
Prior work surveying MTurk workers has identified similar demographic imbalances~\cite{Difallah2018DemographicsAD}.
To correct our demographic skew to be more in line with the U.S. adult population \cite{uscensus}, we augmented the crowd annotators with an additional sample of annotators recruited through social channels and personal networks, focusing on increasing the representation of those who identified as non-White.
These annotators were paid in the form of a \$10 gift card and were assigned one of the conditions randomly.
At the end we recruited 12 additional non-crowd annotators. 

The final demographic distribution of our full annotator pool of 165 participants was: 78.2\% White, 7.9\% Asian, 6.7\% Black, 4.8\% Latino, 0.6\% Native-American, and 1.8\% multiple. The gender distribution of annotators was 40.6\% female, 58.8\% male, 0.6\% non-binary.
We also asked participants to self-report their own skin tone using the annotation tool; the distribution is shown in \autoref{self-skin-tone}.
We discuss limitations of our sample in Section~\ref{limitations}.

\textbf{Designer and Annotator Positionality}:
A limitation of prior literature was the opaqueness of the designer positionality. When the authors did not describe the decision-making behind their process, we could only guess what social, political, and ethical perspectives informed the design. By stating our positionality as designers, we hope to not only reflect on how our identities impacted our design but directly engage with the limits of our perspectives. 
The study was scoped to the U.S. as all of the authors reside in the U.S. and are at U.S. institutions. This informed decisions such as the ethnicity and gender categories we used in our survey.
We also required annotators to be U.S.-based, so that they are also situated in the racial and skin tone-stratification contexts of the U.S. 

For Black populations in the U.S., skin tone has been used since the Atlantic Slave Trade to allocate degrees of social power to enslaved people \cite{monk2021unceasing}. Even today, the associations to skin tone are reflected in sentencing trends in the U.S. justice systems \cite{monk2019color}. Given the context, there is great responsibility placed on an annotator when they select skin tone for themselves and image subjects. Thus, we surveyed annotators about their comfort with the task of annotating skin tone (\autoref{sec:survey}). 
Finally, as mentioned, we asked annotators to mark their own skin tone---this enabled us to investigate another aspect of annotator positionality.

\begin{figure}[t]
    \centering
    \includegraphics[width=\linewidth]{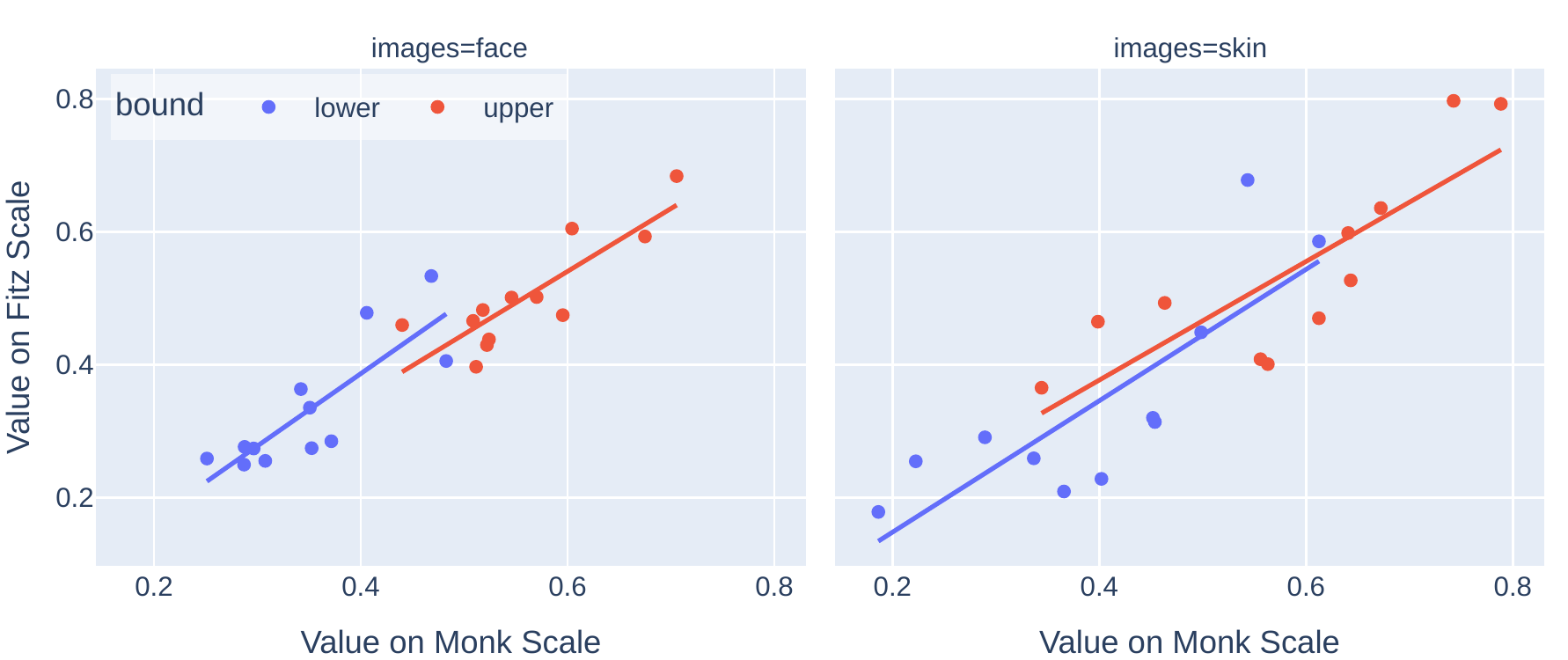}
    \caption{(\textbf{RQ1}) Correlation between the two scale types for both bounds. Each data point indicates the aggregated skin tone rating for the corresponding bound on an image. Higher values indicate darker skin tones. Values produced by \textsc{dl} conditions are remapped onto a light-to-dark scale before aggregating.}
    \label{fig:scale-alignment}
\end{figure}

\subsection{Findings}
\label{sec:findings}

\subsubsection{RQ 1: Do the scales (\textsc{fitz}, \textsc{monk}) correlate with each other for measuring skin tones?}
\label{sec:findings-scale-type}

For our first research question, we wanted to confirm whether there is general agreement in the values produced between the two skin tone annotation scale types we used. Across both image types (\textsc{face} and \textsc{skin}), we found high positive correlation for the upper ($R^2 = 0.721$ and $0.680$ respectively) and lower bounds ($R^2 = 0.692$ and $0.662$ respectively) produced by annotators using the two scales (Figure~\ref{fig:scale-alignment}). 
This result largely serves as a check to validate that both skin tone scales were indeed able to capture differences across a range of skin tones and that annotators were generally able to utilize our annotation interface with existing scales for annotating skin tones. 

\begin{figure}[t]
    \centering
    \includegraphics[width=\linewidth]{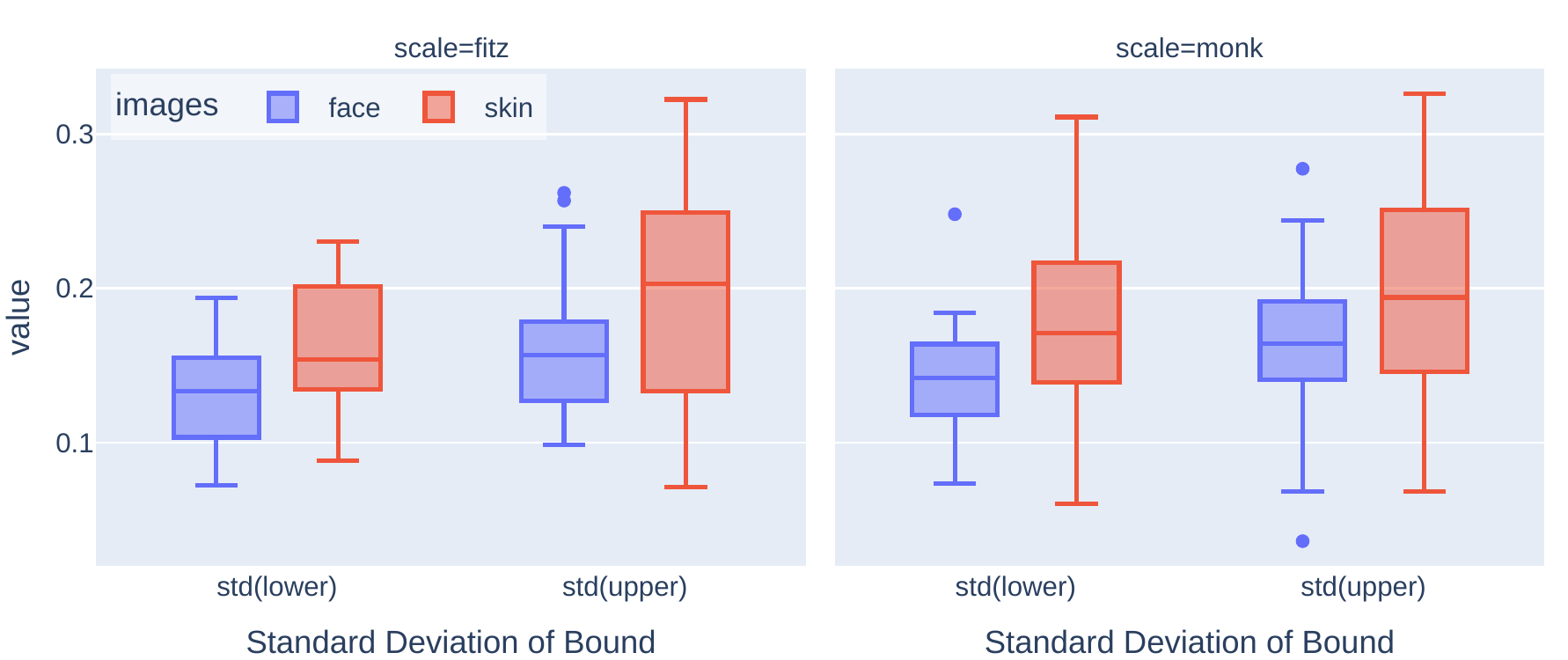}
    \caption{(\textbf{RQ2}) Box plot of agreement (measured as standard deviation of each bound, lower values indicate more agreement) for each scale type and image type. The only significant variable is the \textit{image type} (shown here as red and blue colors). }
    \label{fig:scale-image-type}
\end{figure}

\subsubsection{RQ 2: Does scale type, ordering of scale, or image type affect the agreement between annotators?}
\label{sec:findings-agreement}

To examine whether our control variables affected agreement between annotators, we used a linear model multi-way ANOVA test to compare the effect of the independent variables: scale type (\textsc{fitz}, \textsc{monk}), scale order (\textsc{ld}, \textsc{dl}), and image type (\textsc{face}, \textsc{skin}) variables as well as any pairwise interactions, on the dependent variable of standard deviation of the upper and lower bounds. The standard deviation of each bound is used as it is a common way to characterize inter-annotator agreement in a continuous rating scale setting (Figure~\ref{fig:scale-image-type}).

We found that for both lower and upper bounds, the only significant variable (at $p = 1.5 \times 10^{-3} < 0.05$, and $p = 2.9 \times 10^{-3} < 0.05$ for each bound respectively) that affected annotator agreement was the \textbf{image type}, with images in the \textsc{face} condition showing more agreement than that of \textsc{skin}. As skin tone annotation is most commonly conducted on datasets involving portrait shots with visible faces, this finding may suggest that skin tone annotation on less common image types, like images of skin patches without faces, may result in lower agreement between annotators as they have less context to draw from. However, the inclusion of faces also potentially biases annotators towards using other contextual features such as race and ethnicity, which may have affected their consistency. 

\subsubsection{RQ 3: Does scale type, ordering of scale, or image type affect the uncertainty of each annotator?}
\label{sec:findings-uncertainty}

\begin{figure}[t]
    \includegraphics[width=\linewidth]{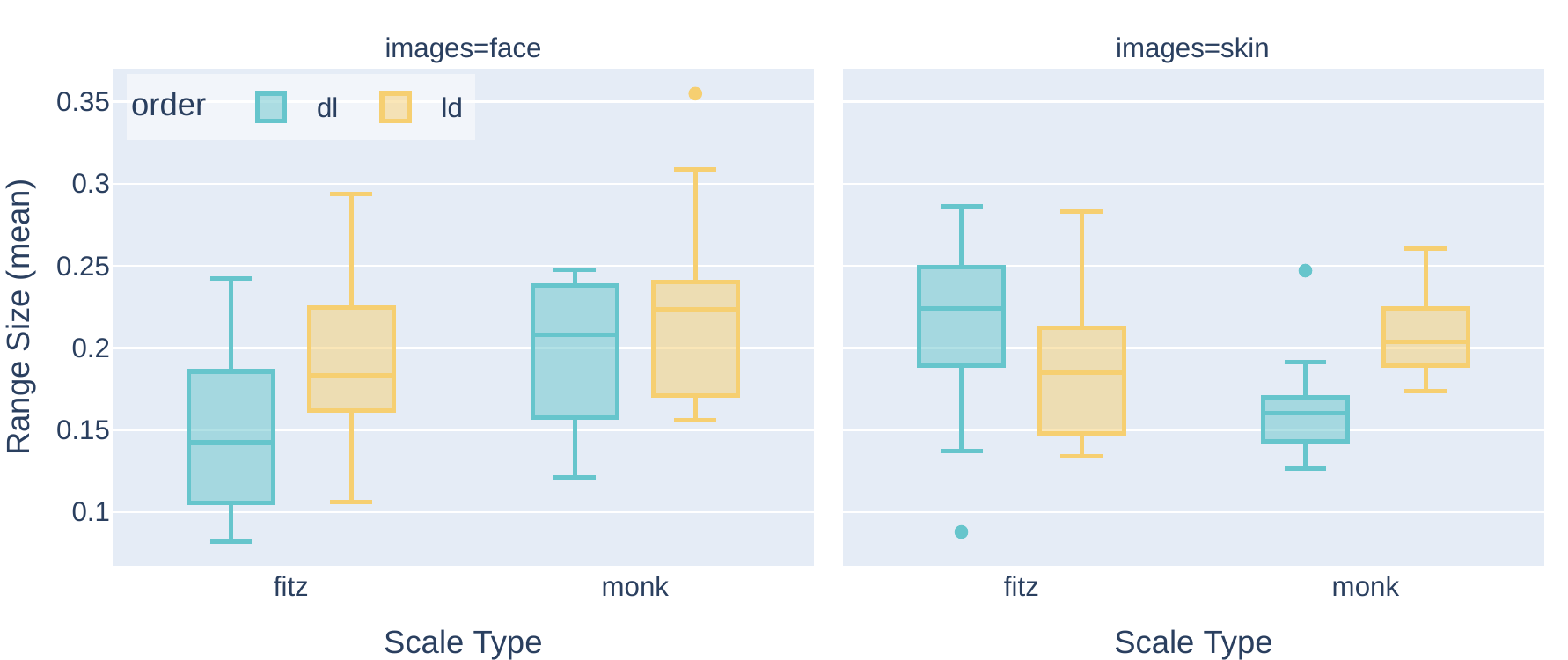}
    \caption{(\textbf{RQ3}) Box plot of uncertainty (measured as the size of each range, lower values indicate more certainty) for each scale type and image type. We see significant effects on uncertainty from both the \textit{scale order} and the \textit{scale type} $\times$ \textit{image type} interaction.}
    \label{fig:corrleate-self}
\end{figure}

While previously we explored the effects of our controlled variables on agreement \textit{between} different annotators, here we examined whether the configuration of annotation system affects each annotator's own \textit{individual} uncertainty during annotation. As we utilized a range-based annotation system, we were able to quantify this uncertainty by examining the \textit{size} of the ranges produced by each annotator. Similar to the section before, we conducted an ANOVA test to compare our independent variables and pairwise interactions against the \textit{size} of the range produced by each annotator (Figure~\ref{fig:corrleate-self}).

We found two significant effects: the \textbf{scale order} ($p = 0.029 < 0.05$) and the interaction of \textbf{scale type} $\times$ \textbf{image type} ($p = 0.010 < 0.05$). Of the latter interaction effect, we found that there was a statistically significant difference between the pairings (\textsc{fitz} x \textsc{face}) and (\textsc{monk} x \textsc{face}) at $p = 0.024 < 0.05$ (identified via Tukey's HSD). Thus, we conclude that the order in which the skin tone is presented can affect individual annotators' own evaluation of their uncertainty. However, \emph{how} it affected uncertainty seems to depend on the type of image and scale being used. Overall, we found that using a \textit{dark-to-light} scale ordering tended to result in lower individual uncertainty. However, when annotating \textsc{skin} images, this trend was not observed for the Fitzpatrick scale. We hypothesize that this may have been the result of multiple factors at work: As a large majority of our annotators self-reported lighter skin tones, when utilizing our annotation process which establishes the lower bound first, a \textit{dark-to-light} scale may result in better estimation of the lower bound from more easily contrasting skin tones, reducing the size of the final range. On the other hand, the lack of additional race and ethnicity context in the \textsc{skin} task may have worked in conjunction with the \textsc{fitz} scale's specialization for directly annotating skin tone on skin patches, reducing this effect.

\begin{figure}[t]
     \centering
     \begin{subfigure}{\linewidth}
        \includegraphics[width=\textwidth]{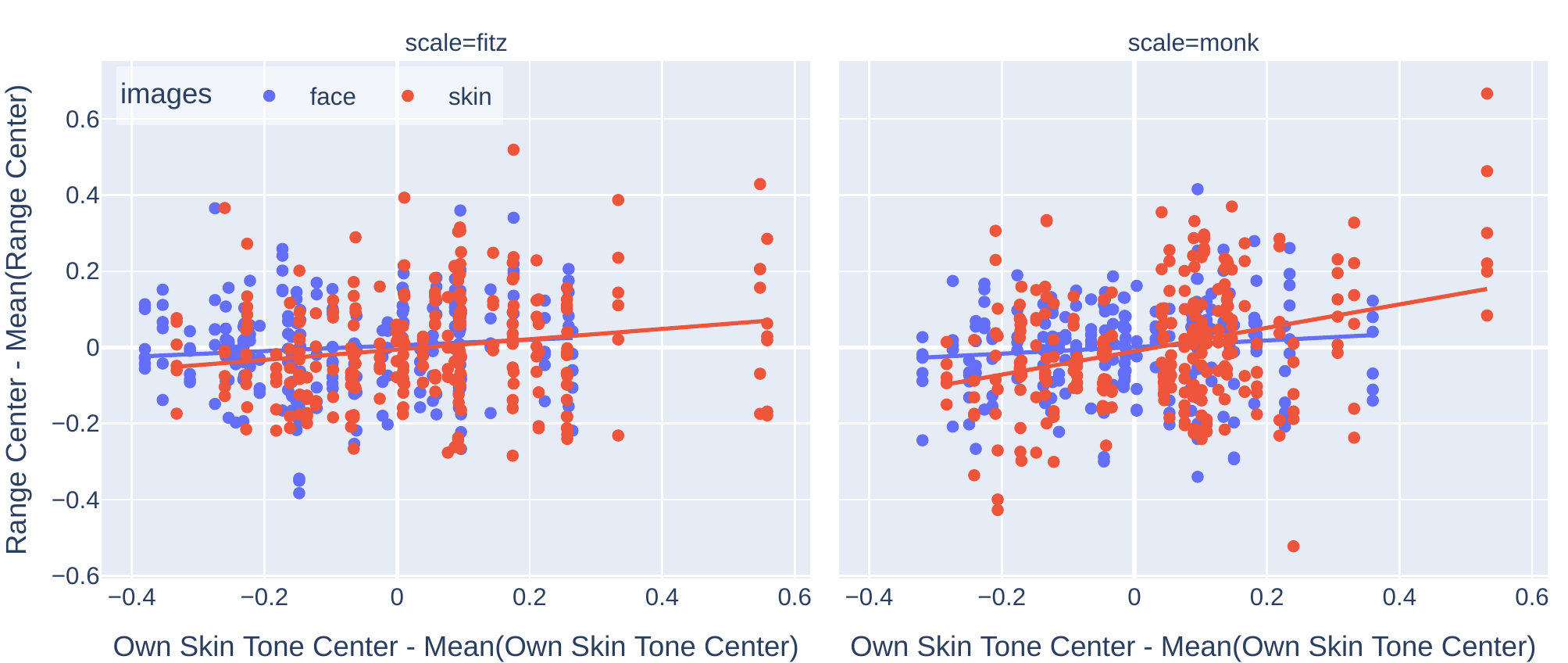}
         \caption{\textbf{Value bias}: x-axis shows the annotator's self-reported skin tone relative to the mean across all annotators. y-axis shows the annotator's judgment of skin tone of each data point relative to the respective mean of that item in the associated condition. Higher = darker skin tone.}
         \label{fig:annot-self-bias:value-bias}
    \end{subfigure}
     \begin{subfigure}{\linewidth}
         \includegraphics[width=\textwidth]{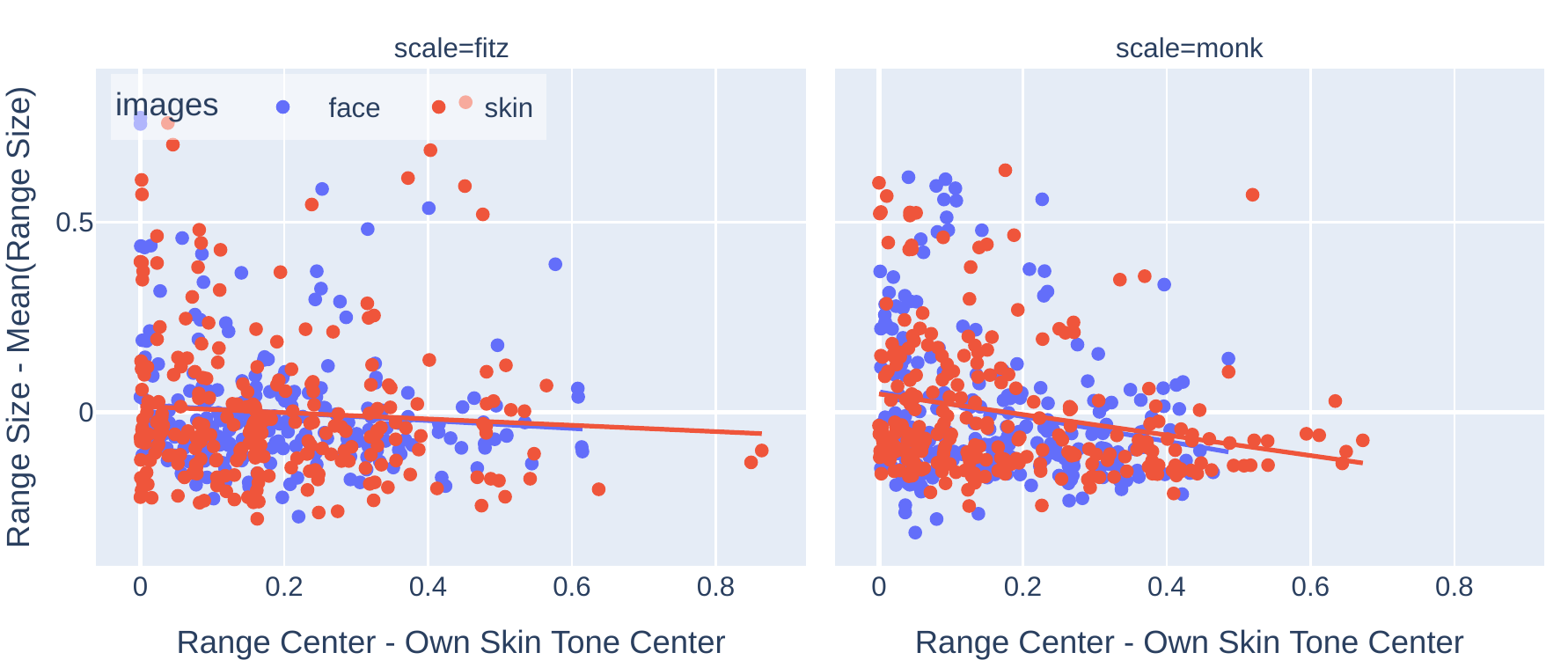}
         \caption{\textbf{Uncertainty bias}: x-axis shows the distance between the annotated item's skin tone and the annotator's self-reported skin tone. y-axis shows the uncertainty of the item relative to the mean uncertainty of that item in the respective condition}
         \label{fig:annot-self-bias:uncertainty-bias}
     \end{subfigure}
     \caption{\textbf{(RQ4}) Two scatter plots showcasing potential correlations between the annotators' own self-reported skin tone and the annotations produced by them.
     }
     \label{fig:annot-self-bias}
\end{figure}

\subsubsection{RQ 4: Does the annotator's own self-reported skin tone (positionality) bias their uncertainty or agreement?}
\label{sec:findings-self}

Finally, we examined whether the annotators' self-reported skin tone correlated with how they produced skin tone annotations. Specifically, we tested for two forms of potential biases that could occur. For \textbf{value bias}, we looked at whether the annotator's own skin tone (relative to the mean across all annotators) correlated with their annotations (relative to the mean across annotations produced under the same experiment conditions). A positive or negative correlation here would indicate that an annotator biases their annotations towards or away from their own skin tone. As shown in Figure~\ref{fig:annot-self-bias:value-bias}, we note a very weak (not significant) positive trend, with the most prominent $R^2 = 0.100$ observed for \textsc{skin} images using the \textsc{monk} scale. This suggests that annotators may potentially bias their annotated skin tone towards their own. However, we also note the caveat that data points get much sparser the further we get from the mean skin tone, which we hypothesize is likely due to the demographic concentration of our annotators.

For \textbf{uncertainty bias}, we examined whether the difference between the annotated image's skin tone and the annotator's own skin tone correlated with their self-reported uncertainty in the form of the size of their range. 
A positive or negative correlation here would indicate that an annotator is more or less uncertain the further the annotated image's skin tone is from their own skin tone.
As shown in Figure~\ref{fig:annot-self-bias:uncertainty-bias}, we note a very weak (not significant) negative trend, with the most prominent $R^2 = 0.056$ observed for \textsc{skin} images using the \textsc{monk} scale. This may suggest that annotators are potentially more certain when annotating images that have a skin tone different from their own. However, as before, any potential trends may be a result of the lack of annotators on the darker end of the skin tone spectrum.

%% file: 005_discussion.tex
\section{Discussion and Recommendations}

\textbf{Associations and Differences between Skin Tone and Race}.
The degree to which race data is embedded in the skin tone annotation process often reveals a Western hegemonic understanding of the association of skin tone to race. 
While we indeed observed correlations between self-reported skin tone and race in our annotation study, such correlations are far from a perfect association between the two.
We argue that skin tone annotations are not just a proxy to associate certain ethnic groups to skin tone categories; instead they capture a spectrum of additional complexions not afforded by simple ethnic groups and race demographics. 
This can be especially salient for underrepresented populations within the Black and Asian diasporas. 
In papers that have noted the lack of representation of skin tones of Black or Asian populations \cite{ware2020racial}, authors have suggested including skin tones to better represent the extent of possible skin tones among these populations. 
Throughout the literature review, we observed implicit and explicit associations between skin tone and race that manifested in the designers' decision-making.
We encourage researchers seeking to engage with datasets involving skin tone and race to take care in capturing these associations. For instance, they could capture annotator positionality through annotator self-reported skin tone and demographic information in conjunction with the collection of annotations.

\textbf{Uncertainty Present in Skin Tone Annotations}.
In both our literature review and our own experiments, we have observed that even the more objective measurement of skin tone  is still associated with factors of uncertainty. 
Within our study, aspects like the ordering of the scale were observed to affect individual uncertainty, while the type of image affected collective agreement. 
In practice, the aspect of uncertainty in annotations can often be overlooked, leading to over-estimates of the capabilities of downstream models~\cite{gordon2021-disagreementdeconv}.
While design factors of our study limited our ability to further examine the mechanisms behind the observed uncertainty, our results demonstrate the need for recruiting a diverse pool of annotators when conducting skin tone annotations and collecting multiple annotations per item so that uncertainty can be identified.
We also urge anyone collecting skin tone annotations for modeling to be cognizant about how uncertainty may limit the actual capabilities of models produced. 

\textbf{Transparency in Skin Tone Annotation Procedures}.
As noted in our literature review, more than 60\% of the corpus had ambiguously documented procedures. The lack of transparency and consistent reporting in skin tone annotation processes is a major issue for current work using skin tone annotation. 
In addition, our review highlighted components of the annotation process that could be standardized or at least explained in a more meaningful way.
Coupled with our findings of factors that affected annotation results in the annotation study, we believe that providing transparency around the procedures and configuration decisions in skin tone annotation will be an important step if we are to use skin tone data as ground truth for ensuring the fairness of algorithms and datasets.
Indeed, even the most robust skin tone annotation procedures in prior work reported some level of unreliability in their annotations, meaning that transparency will be an important way for downstream users to understand the limitations resulting from the subjectivity of skin tone annotation. 

\textbf{Consider Effects from Scale Order}.
Finally, in our annotation experiments, we found that arranging the scale from \textit{darker to lighter} somewhat surprisingly correlated with a significant reduction in individual annotators' uncertainty. 
The hegemony of lighter to darker hierarchy associated to skin tone stratification might have played a role in creating implicit biases leading to this trend.
For example, FST is encoded with category values that increase in value from the lighter to darker skin tones. 
In the literature review, even when the scale was not FST, scale values were often ordered from lighter to darker categories. 
In a society in which lighter skin tones yield a social proximity to whiteness, ordering skin tone with lighter skin first enforces the hierarchy of social power. 
Since our tool establishes the lower boundary first, starting with a skin tone spectrum default of lighter to darker shades could mean that our set of majority-White U.S.-based annotators produce a less precise rating when exploring the spectrum of lighter skin tones.
On the other hand, when the annotator is presented the less common darker-to-lighter scale, the novelty may have encouraged the annotators to consider their annotations with more intention.
We pose that the interaction between how scales are explored and ordered should be an aspect to which researchers collecting skin tone annotations pay greater attention.

\textbf{Consider Effects from Dataset Type}.
Another area where we identified sensitivity was that of the type of image data involved in the annotation. 
In our own experiments, the \textsc{skin} dataset involved the classification of skin tones based on images of skin disease rather than the more common classification of portrait photos.
Some annotators expressed in their free-response feedback that they were caught off guard by the unsettling nature of such images even though their nature was indicated in our recruitment, task prompts, and consent forms. Given this, we recommend taking additional care around the task design and training phases when conducting annotations on potentially sensitive image types.

More generally though, our experiments exposed the sensitivity of skin tone annotation processes to the types of images being annotated. 
In our results in Section~\ref{sec:findings-agreement}, we hypothesized that the additional context provided through the portraits relating to the subject's race and ethnicity may be the reason higher agreement is observed. 
Krishnapriya et al.~\cite{krishnapriya2022analysis} went so far as to suggest that human annotators were more consistent than the fully automated system because they factored in the race of the image subjects in their skin tone analysis.

\section{Limitations and Future Work}
\label{limitations}
One limitation of our study arises from the scope of the annotation task and the recruitment of annotators.
While our study explored whether potential differences exist between different image types, we only utilized a limited sub-sample of images of each type and measured along the two most commonly used skin tone scales.
A larger scale study involving a larger set of images that control for additional factors would be helpful in shedding light on \textit{how} different factors affect annotations rather than our exploration of \textit{whether} they do.

Additionally, a majority of our annotators were recruited through online crowdsourcing platforms, leading to an observed imbalance in the annotator demographics (Section \ref{sec:method}).
This is not surprising, as online crowd work platforms have long been known to have a skewed demographic breakdown compared to the general population, presenting challenges for studies involving subjectivity~\cite{Difallah2018DemographicsAD,Kazai2012TheFO,Ross2010WhoAT}. 
Even though our additional recruitment yielded a more representative final demographic distribution, the majority of annotators overall still identified as White, contributing to an over-representation of lighter skin tones in self-reported ratings (\autoref{self-skin-tone}).
In addition, we were not able to recruit subject-matter experts (i.e., dermatology) to provide an expert-informed source for us to evaluate how differences we observed impacted the final quality or correctness of annotations themselves in relation to scales like FST.
We note that works in our literature review have used dermatologists to validate annotations~\cite{buolamwini2018gender} or provide a set of annotations as a contribution to inter-rater uncertainty analyses and consensus scores~\cite{groh2022towards}.

Finally, we did not compare our manual annotations against fully automated skin tone evaluation metrics and systems based on pixel data. 
While human-centric skin tone data and scales (like FST and MST) remain widely used, there is a body of work that proposes to use automated metrics and systems to establish skin tone as a more objectively-defined concept. 
As the act of annotation itself is a human and socially-situated activity, exploring similarities and differences between automation-centered metrics and systems versus human-defined skin tone judgments is an interesting avenue for future work. 

\section{Conclusion}
The central goal of this work was to investigate the social subjectivity inherent in skin tone annotations for computer vision evaluation. We achieved this through cataloguing how subjectivity was addressed in prior annotation processes, as well as through experimentation. 
Our literature review showed great variability in annotation procedures, an overall lack of documentation, and a lack of engagement with subjectivity.
Our experiment findings indicated that many factors, like the type of data and the configuration of the annotation process, can affect aspects like agreement and uncertainty around the data produced by skin tone annotation.
We contribute to the effort to address disparities in CV models by investigating how factors related to the data, annotators, and annotation process can all impact skin tone annotation outcomes.
We call upon the broader community to commit to moving towards more robust, principled, and well-documented processes when working with CV tasks involving skin tones.

%% file: 010_appendix.tex
\appendix

\newpage
\onecolumn

\section{Literature Review Dimensions of Analysis}
\label{dimensions-analysis}

Below in \autoref{review_def}, we list all dimensions the first author used to qualitatively analyze the papers in our literature review. We also provide a more detailed definition of that dimension along with an example of an annotation for a paper. The full dataset of papers along with the annotated values for each dimension is provided at \url{https://github.com/Social-Futures-Lab/skin-deep/}.

\begin{table*}[ht]
\small
\begin{center}
\begin{tabular}{p{0.25\linewidth}|p{0.4\linewidth}|p{0.3\linewidth}}
 \toprule
 \textbf{Dimension} & \textbf{Definition} & \textbf{Examples} \\
 \bottomrule
 \toprule
 \multicolumn{3}{c}{Dataset Information} \\
 \midrule
 Task & The intended Computer Vision task that will be trained or tested by the annotated dataset. & Facial Recognition\\
 \hline
 Year & The year the paper was published.  & 2017 \\
 \hline
 Dataset Source & The source of the images or videos in the annotated dataset. & Flickr, CelebA \\
 \hline
 Publicly Available & The resulting dataset (including the annotations) is available to the public & Yes/No \\
 \hline
 No. of Subjects & The number of individual human subjects represented in the dataset & 15, 20\\
 \hline
 No. of Images/Videos & The number of images or videos are a part of the dataset & 100, 40/300, 40 videos\\
 \bottomrule
 \toprule
 \multicolumn{3}{c}{Annotation Procedure} \\
 \midrule
 Procedure Description & An explanation of the skin tone scale definition, annotation platform and procedural restrictions & ``The apparent skin tone and lighting attributes were labeled by a group of human evaluators \cite{menezes2021bias}.''  \\
 \hline
 Annotator & The individual or individuals who conducted the skin tone annotations. & Author(s), Experts\\
 \hline
 Scale/Measurement & The numerical scale or categories used to describe the range of skin tones represented by the annotations. & Fitzpatrick scale, ``lighter'' and ``darker'' \\
 \hline
 Annotation Distribution & The proportion of each skin tone category present in the annotated dataset. & Lighter (50\%), Darker (50\%) \\
 \bottomrule
 \toprule
 \multicolumn{3}{c}{Annotation Uncertainty} \\
 \midrule
 Analysis Description & An explanation of the methods,  calculations, or qualitative findings used to evaluate inter-rater agreement or uncertainty & ``We compute the Cohen’s k-coefficient for
all pairs of participants with more than 5 common images in their surveys \cite{celis2020implicit}.'' \\
 \hline
 Analysis Results & Any quantitative metric for the uncertainty of the skin tone annotations. & ``k-coefficient: 0.58 (median: 0.62)'' \\
 \bottomrule
 \toprule
 \multicolumn{3}{c}{Race Annotations or Metadata} \\
 \midrule
 Racial Categories & The categories used to describe the various races or ethnicities of the dataset subjects. & Black, South Asian, Western European \\
 \hline
 Category Distribution & The proportion of each racial category presented in the annotated dataset. & Black (100\%) \\
 \bottomrule
 \toprule
 \multicolumn{3}{c}{Additional Information} \\
 \midrule
 Comments & Relevant excerpts, thoughts or external work that provide context to the annotation process in the work. & The work also used a dataset that was already annotated for skin tone. \\
 \bottomrule
\end{tabular}
\caption{The definitions and examples of all the dimensions evaluated in the literature review.}
\label{review_def}
\end{center}
\end{table*}

\newpage

\section{Annotator Self-Reported Skin Tone Distributions}
\label{self-skin-tone}
\begin{figure}[ht!]
    \centering
    \includegraphics[width=\textwidth]{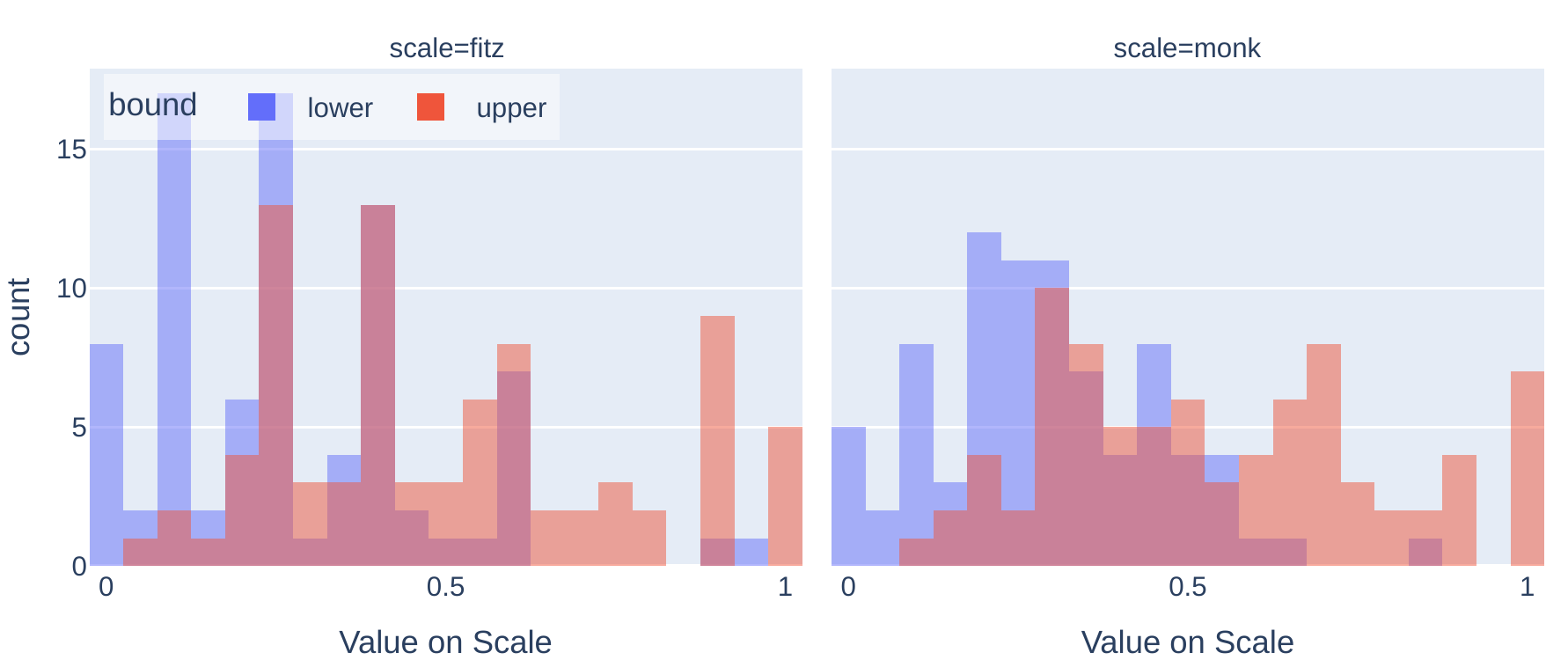}
    \caption{A histogram of the distribution of self-reported skin tone values (upper and lower bounds) shown for each scale. Lower values indicate lighter skin tone while higher values indicate darker.}
    \label{fig:self-skin-tone}
\end{figure}

We see a diverse spread of skin tone values as self reported by annotators, with a noted bias towards the lighter side of each scale overall (\autoref{fig:self-skin-tone}). Skin tone values can present a space that is much richer than simple racial demographics. We can also observe differences between the two scales, where, despite the entire continuous range [0, 1] being available in both cases, the less-granular \textsc{fitz} scale results in less resolution compared to the \textsc{monk} scale that contains more levels.

\section{Post-Annotation Survey}
\label{sec:survey}
After the main annotation task the participants were prompted to answer the following statements with likert scale response of agreement (strongly disagree, disagree, neutral, agree, strongly agree):

\begin{itemize}
    \item I am confident in the labeling of my skin tone.
    \item I am confident in my annotations of the images.
    \item Throughout history people have been categorized by the color of their skin by other individuals placed in positions of social power. These categorizations were used to enforce harmful social hierarchies based on racism and colorism (i.e. \href{https://allthatsinteresting.com/brown-paper-bag-test}{Jim Crow era in Southern US}). Even today in the United States, people have different life experiences based on the long-term impact of skin tone categorizations (i.e. \href{https://www.tandfonline.com/doi/abs/10.1080/01419870.2018.1508736?scroll=top&needAccess=true&journalCode=rers20}{arrest rates in the United States}). Given this particular context, indicate how much you agree with the following statements:
    
    I considered the race and/or ethnicity of the image subject while annotating.
    \item I felt comfortable annotating the skin tone of the given images in this task.
    \item I felt comfortable annotating my \textbf{own} skin tone.
\end{itemize}

To further clarify the statements, short definitions of key words were provided as well.
\begin{itemize}
    \item Confident: certain to a significant degree
    \item Consider: take into account, especially before making a decision
\end{itemize}

\subsection{Results from Annotator Self-Reported Experiences}
\label{sec:findings-reported-experience}
\begin{figure}[ht!]
    \centering
    \includegraphics[width=\textwidth]{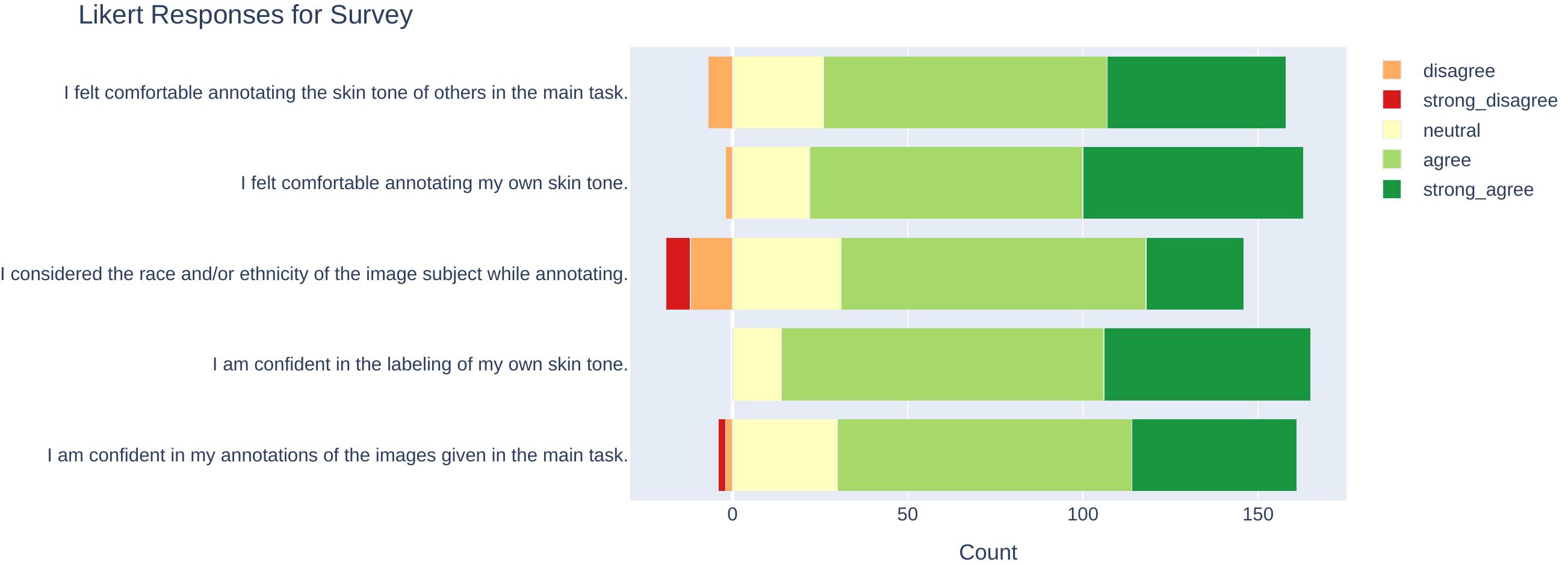}
    \caption{Diverging stacked bar chart of responses to the survey.}     
    \label{fig:self-report}
\end{figure}

We examined the survey results to explore how the participants perceived the task of annotating skin tone. As can be seen in the results from Figure~\ref{fig:self-report}, overall most annotators were comfortable with the idea of annotating skin tone in general (both for the images and themselves). We also observed that compared to the other questions, more annotators self reported disagreement with the statement that race and/or ethnicity were a factor in their annotation consideration.

%% file: main.bbl

\begin{thebibliography}{89}


\ifx \showCODEN    \undefined \def \showCODEN     #1{\unskip}     \fi
\ifx \showDOI      \undefined \def \showDOI       #1{#1}\fi
\ifx \showISBNx    \undefined \def \showISBNx     #1{\unskip}     \fi
\ifx \showISBNxiii \undefined \def \showISBNxiii  #1{\unskip}     \fi
\ifx \showISSN     \undefined \def \showISSN      #1{\unskip}     \fi
\ifx \showLCCN     \undefined \def \showLCCN      #1{\unskip}     \fi
\ifx \shownote     \undefined \def \shownote      #1{#1}          \fi
\ifx \showarticletitle \undefined \def \showarticletitle #1{#1}   \fi
\ifx \showURL      \undefined \def \showURL       {\relax}        \fi
\providecommand\bibfield[2]{#2}
\providecommand\bibinfo[2]{#2}
\providecommand\natexlab[1]{#1}
\providecommand\showeprint[2][]{arXiv:#2}

\bibitem[Al-Naji and Chahl(2017)]%
        {al2017simultaneous}
\bibfield{author}{\bibinfo{person}{Ali Al-Naji} {and} \bibinfo{person}{Javaan
  Chahl}.} \bibinfo{year}{2017}\natexlab{}.
\newblock \showarticletitle{Simultaneous tracking of cardiorespiratory signals
  for multiple persons using a machine vision system with noise artifact
  removal}.
\newblock \bibinfo{journal}{\emph{IEEE journal of translational engineering in
  health and medicine}}  \bibinfo{volume}{5} (\bibinfo{year}{2017}),
  \bibinfo{pages}{1--10}.
\newblock


\bibitem[Al-Naji and Chahl(2018)]%
        {al2018remote}
\bibfield{author}{\bibinfo{person}{Ali Al-Naji} {and} \bibinfo{person}{Javaan
  Chahl}.} \bibinfo{year}{2018}\natexlab{}.
\newblock \showarticletitle{Remote optical cardiopulmonary signal extraction
  with noise artifact removal, multiple subject detection \& long-distance}.
\newblock \bibinfo{journal}{\emph{IEEE Access}}  \bibinfo{volume}{6}
  (\bibinfo{year}{2018}), \bibinfo{pages}{11573--11595}.
\newblock


\bibitem[Aledhari et~al\mbox{.}(2021)]%
        {aledhari2021multimodal}
\bibfield{author}{\bibinfo{person}{Mohammed Aledhari}, \bibinfo{person}{Rehma
  Razzak}, \bibinfo{person}{Reza~M Parizi}, {and} \bibinfo{person}{Gautam
  Srivastava}.} \bibinfo{year}{2021}\natexlab{}.
\newblock \showarticletitle{Multimodal machine learning for pedestrian
  detection}. In \bibinfo{booktitle}{\emph{2021 IEEE 93rd Vehicular Technology
  Conference (VTC2021-Spring)}}. IEEE, \bibinfo{pages}{1--7}.
\newblock


\bibitem[Almeida et~al\mbox{.}(2022)]%
        {almeida2022ethics}
\bibfield{author}{\bibinfo{person}{Denise Almeida}, \bibinfo{person}{Konstantin
  Shmarko}, {and} \bibinfo{person}{Elizabeth Lomas}.}
  \bibinfo{year}{2022}\natexlab{}.
\newblock \showarticletitle{The ethics of facial recognition technologies,
  surveillance, and accountability in an age of artificial intelligence: a
  comparative analysis of US, EU, and UK regulatory frameworks}.
\newblock \bibinfo{journal}{\emph{AI and Ethics}} \bibinfo{volume}{2},
  \bibinfo{number}{3} (\bibinfo{year}{2022}), \bibinfo{pages}{377--387}.
\newblock


\bibitem[Anderson et~al\mbox{.}(2019)]%
        {anderson2019understanding}
\bibfield{author}{\bibinfo{person}{Janet Anderson}, \bibinfo{person}{Charles
  Otto}, \bibinfo{person}{Brianna Maze}, \bibinfo{person}{Nathan Kalka}, {and}
  \bibinfo{person}{James~A Duncan}.} \bibinfo{year}{2019}\natexlab{}.
\newblock \showarticletitle{Understanding confounding factors in face detection
  and recognition}. In \bibinfo{booktitle}{\emph{2019 International Conference
  on Biometrics (ICB)}}. IEEE, \bibinfo{pages}{1--8}.
\newblock


\bibitem[Arosarena(2015)]%
        {arosarena2015options}
\bibfield{author}{\bibinfo{person}{Oneida Arosarena}.}
  \bibinfo{year}{2015}\natexlab{}.
\newblock \showarticletitle{Options and challenges for facial rejuvenation in
  patients with higher fitzpatrick skin phototypes}.
\newblock \bibinfo{journal}{\emph{JAMA Facial Plastic Surgery}}
  \bibinfo{volume}{17}, \bibinfo{number}{5} (\bibinfo{year}{2015}),
  \bibinfo{pages}{358--359}.
\newblock


\bibitem[Attia and Edge(2017)]%
        {attia2017ing}
\bibfield{author}{\bibinfo{person}{Mariam Attia} {and} \bibinfo{person}{Julian
  Edge}.} \bibinfo{year}{2017}\natexlab{}.
\newblock \showarticletitle{Be (com) ing a reflexive researcher: a
  developmental approach to research methodology}.
\newblock \bibinfo{journal}{\emph{Open Review of Educational Research}}
  \bibinfo{volume}{4}, \bibinfo{number}{1} (\bibinfo{year}{2017}),
  \bibinfo{pages}{33--45}.
\newblock


\bibitem[Bagdasaryan et~al\mbox{.}(2019)]%
        {bagdasaryan2019differential}
\bibfield{author}{\bibinfo{person}{Eugene Bagdasaryan}, \bibinfo{person}{Omid
  Poursaeed}, {and} \bibinfo{person}{Vitaly Shmatikov}.}
  \bibinfo{year}{2019}\natexlab{}.
\newblock \showarticletitle{Differential privacy has disparate impact on model
  accuracy}.
\newblock \bibinfo{journal}{\emph{Advances in neural information processing
  systems}}  \bibinfo{volume}{32} (\bibinfo{year}{2019}).
\newblock


\bibitem[Bahmani et~al\mbox{.}(2021)]%
        {bahmani2021sreds}
\bibfield{author}{\bibinfo{person}{Keivan Bahmani}, \bibinfo{person}{Richard
  Plesh}, \bibinfo{person}{Chinmay Sahu}, \bibinfo{person}{Mahesh Banavar},
  {and} \bibinfo{person}{Stephanie Schuckers}.}
  \bibinfo{year}{2021}\natexlab{}.
\newblock \showarticletitle{SREDS: A dichromatic separation based measure of
  skin color}. In \bibinfo{booktitle}{\emph{2021 IEEE International Workshop on
  Biometrics and Forensics (IWBF)}}. IEEE, \bibinfo{pages}{1--6}.
\newblock


\bibitem[Beal et~al\mbox{.}(2022)]%
        {beal2022billion}
\bibfield{author}{\bibinfo{person}{Josh Beal}, \bibinfo{person}{Hao-Yu Wu},
  \bibinfo{person}{Dong~Huk Park}, \bibinfo{person}{Andrew Zhai}, {and}
  \bibinfo{person}{Dmitry Kislyuk}.} \bibinfo{year}{2022}\natexlab{}.
\newblock \showarticletitle{Billion-scale pretraining with vision transformers
  for multi-task visual representations}. In
  \bibinfo{booktitle}{\emph{Proceedings of the IEEE/CVF Winter Conference on
  Applications of Computer Vision}}. \bibinfo{pages}{564--573}.
\newblock


\bibitem[Birhane(2021)]%
        {birhane2021algorithmic}
\bibfield{author}{\bibinfo{person}{Abeba Birhane}.}
  \bibinfo{year}{2021}\natexlab{}.
\newblock \showarticletitle{Algorithmic injustice: a relational ethics
  approach}.
\newblock \bibinfo{journal}{\emph{Patterns}} \bibinfo{volume}{2},
  \bibinfo{number}{2} (\bibinfo{year}{2021}), \bibinfo{pages}{100205}.
\newblock


\bibitem[Buolamwini and Gebru(2018)]%
        {buolamwini2018gender}
\bibfield{author}{\bibinfo{person}{Joy Buolamwini} {and}
  \bibinfo{person}{Timnit Gebru}.} \bibinfo{year}{2018}\natexlab{}.
\newblock \showarticletitle{Gender shades: Intersectional accuracy disparities
  in commercial gender classification}. In \bibinfo{booktitle}{\emph{Conference
  on fairness, accountability and transparency}}. PMLR,
  \bibinfo{pages}{77--91}.
\newblock


\bibitem[Bureau(2020)]%
        {uscensus}
\bibfield{author}{\bibinfo{person}{United States~Census Bureau}.}
  \bibinfo{year}{2020}\natexlab{}.
\newblock \bibinfo{title}{2020 Census Results}.
\newblock
\newblock
\newblock
\shownote{\url{https://www.census.gov/programs-surveys/decennial-census/decade/2020/2020-census-results.html}}.


\bibitem[Celis and Keswani(2020)]%
        {celis2020implicit}
\bibfield{author}{\bibinfo{person}{L~Elisa Celis} {and} \bibinfo{person}{Vijay
  Keswani}.} \bibinfo{year}{2020}\natexlab{}.
\newblock \showarticletitle{Implicit diversity in image summarization}.
\newblock \bibinfo{journal}{\emph{Proceedings of the ACM on Human-Computer
  Interaction}} \bibinfo{volume}{4}, \bibinfo{number}{CSCW2}
  (\bibinfo{year}{2020}), \bibinfo{pages}{1--28}.
\newblock


\bibitem[Chang et~al\mbox{.}(2018)]%
        {chang2018robust}
\bibfield{author}{\bibinfo{person}{Cheng-Chun Chang}, \bibinfo{person}{Shi-Tien
  Hsing}, \bibinfo{person}{Yung-Chi Chuang}, \bibinfo{person}{Chien-Ta Wu},
  \bibinfo{person}{Tung-Jing Fang}, \bibinfo{person}{Kuan-Fu Chen}, {and}
  \bibinfo{person}{Bill Choi}.} \bibinfo{year}{2018}\natexlab{}.
\newblock \showarticletitle{Robust skin type classification using convolutional
  neural networks}. In \bibinfo{booktitle}{\emph{2018 13th IEEE Conference on
  Industrial Electronics and Applications (ICIEA)}}. IEEE,
  \bibinfo{pages}{2011--2014}.
\newblock


\bibitem[Chardon et~al\mbox{.}(1991)]%
        {chardon1991skin}
\bibfield{author}{\bibinfo{person}{Alain Chardon}, \bibinfo{person}{Isabelle
  Cretois}, {and} \bibinfo{person}{Colette Hourseau}.}
  \bibinfo{year}{1991}\natexlab{}.
\newblock \showarticletitle{Skin colour typology and suntanning pathways}.
\newblock \bibinfo{journal}{\emph{International journal of cosmetic science}}
  \bibinfo{volume}{13}, \bibinfo{number}{4} (\bibinfo{year}{1991}),
  \bibinfo{pages}{191--208}.
\newblock


\bibitem[Chen et~al\mbox{.}(2021)]%
        {chen2021goldilocks}
\bibfield{author}{\bibinfo{person}{Quan~Ze Chen}, \bibinfo{person}{Daniel~S
  Weld}, {and} \bibinfo{person}{Amy~X Zhang}.} \bibinfo{year}{2021}\natexlab{}.
\newblock \showarticletitle{Goldilocks: Consistent crowdsourced scalar
  annotations with relative uncertainty}.
\newblock \bibinfo{journal}{\emph{Proceedings of the ACM on Human-Computer
  Interaction}} \bibinfo{volume}{5}, \bibinfo{number}{CSCW2}
  (\bibinfo{year}{2021}), \bibinfo{pages}{1--25}.
\newblock


\bibitem[Cook et~al\mbox{.}(2019)]%
        {cook2019demographic}
\bibfield{author}{\bibinfo{person}{Cynthia~M Cook}, \bibinfo{person}{John~J
  Howard}, \bibinfo{person}{Yevgeniy~B Sirotin}, \bibinfo{person}{Jerry~L
  Tipton}, {and} \bibinfo{person}{Arun~R Vemury}.}
  \bibinfo{year}{2019}\natexlab{}.
\newblock \showarticletitle{Demographic effects in facial recognition and their
  dependence on image acquisition: An evaluation of eleven commercial systems}.
\newblock \bibinfo{journal}{\emph{IEEE Transactions on Biometrics, Behavior,
  and Identity Science}} \bibinfo{volume}{1}, \bibinfo{number}{1}
  (\bibinfo{year}{2019}), \bibinfo{pages}{32--41}.
\newblock


\bibitem[Costanza-Chock et~al\mbox{.}(2022)]%
        {costanza2022audits}
\bibfield{author}{\bibinfo{person}{Sasha Costanza-Chock},
  \bibinfo{person}{Inioluwa~Deborah Raji}, {and} \bibinfo{person}{Joy
  Buolamwini}.} \bibinfo{year}{2022}\natexlab{}.
\newblock \showarticletitle{Who Audits the Auditors? Recommendations from a
  field scan of the algorithmic auditing ecosystem}. In
  \bibinfo{booktitle}{\emph{2022 ACM Conference on Fairness, Accountability,
  and Transparency}}. \bibinfo{pages}{1571--1583}.
\newblock


\bibitem[Daneshjou et~al\mbox{.}(2022)]%
        {daneshjou2022disparities}
\bibfield{author}{\bibinfo{person}{Roxana Daneshjou}, \bibinfo{person}{Kailas
  Vodrahalli}, \bibinfo{person}{Roberto~A Novoa}, \bibinfo{person}{Melissa
  Jenkins}, \bibinfo{person}{Weixin Liang}, \bibinfo{person}{Veronica
  Rotemberg}, \bibinfo{person}{Justin Ko}, \bibinfo{person}{Susan~M Swetter},
  \bibinfo{person}{Elizabeth~E Bailey}, \bibinfo{person}{Olivier Gevaert},
  {et~al\mbox{.}}} \bibinfo{year}{2022}\natexlab{}.
\newblock \showarticletitle{Disparities in dermatology AI performance on a
  diverse, curated clinical image set}.
\newblock \bibinfo{journal}{\emph{Science advances}} \bibinfo{volume}{8},
  \bibinfo{number}{31} (\bibinfo{year}{2022}), \bibinfo{pages}{eabq6147}.
\newblock


\bibitem[Del~Bino et~al\mbox{.}(2006)]%
        {del2006relationship}
\bibfield{author}{\bibinfo{person}{S Del~Bino}, \bibinfo{person}{J Sok},
  \bibinfo{person}{E Bessac}, {and} \bibinfo{person}{F Bernerd}.}
  \bibinfo{year}{2006}\natexlab{}.
\newblock \showarticletitle{Relationship between skin response to ultraviolet
  exposure and skin color type}.
\newblock \bibinfo{journal}{\emph{Pigment cell research}} \bibinfo{volume}{19},
  \bibinfo{number}{6} (\bibinfo{year}{2006}), \bibinfo{pages}{606--614}.
\newblock


\bibitem[Denton et~al\mbox{.}(2021)]%
        {denton2021whose}
\bibfield{author}{\bibinfo{person}{Emily Denton}, \bibinfo{person}{Mark
  D{\'\i}az}, \bibinfo{person}{Ian Kivlichan}, \bibinfo{person}{Vinodkumar
  Prabhakaran}, {and} \bibinfo{person}{Rachel Rosen}.}
  \bibinfo{year}{2021}\natexlab{}.
\newblock \showarticletitle{Whose ground truth? accounting for individual and
  collective identities underlying dataset annotation}.
\newblock \bibinfo{journal}{\emph{arXiv preprint arXiv:2112.04554}}
  (\bibinfo{year}{2021}).
\newblock


\bibitem[Denton et~al\mbox{.}(2020)]%
        {denton2020bringing}
\bibfield{author}{\bibinfo{person}{Emily Denton}, \bibinfo{person}{Alex Hanna},
  \bibinfo{person}{Razvan Amironesei}, \bibinfo{person}{Andrew Smart},
  \bibinfo{person}{Hilary Nicole}, {and} \bibinfo{person}{Morgan~Klaus
  Scheuerman}.} \bibinfo{year}{2020}\natexlab{}.
\newblock \showarticletitle{Bringing the people back in: Contesting benchmark
  machine learning datasets}.
\newblock \bibinfo{journal}{\emph{arXiv preprint arXiv:2007.07399}}
  (\bibinfo{year}{2020}).
\newblock


\bibitem[Difallah et~al\mbox{.}(2018)]%
        {Difallah2018DemographicsAD}
\bibfield{author}{\bibinfo{person}{Djellel~Eddine Difallah},
  \bibinfo{person}{Elena Filatova}, {and} \bibinfo{person}{Panagiotis~G.
  Ipeirotis}.} \bibinfo{year}{2018}\natexlab{}.
\newblock \showarticletitle{Demographics and Dynamics of Mechanical Turk
  Workers}.
\newblock \bibinfo{journal}{\emph{Proceedings of the Eleventh ACM International
  Conference on Web Search and Data Mining}} (\bibinfo{year}{2018}).
\newblock


\bibitem[Doshi et~al\mbox{.}(2022)]%
        {doshi2022fhp}
\bibfield{author}{\bibinfo{person}{Manan Doshi}, \bibinfo{person}{Jimil Shah},
  \bibinfo{person}{Rahul Soni}, {and} \bibinfo{person}{Soni Bhambar}.}
  \bibinfo{year}{2022}\natexlab{}.
\newblock \showarticletitle{FHP: Facial and Hair Feature Processor for
  Hairstyle Recommendation}. In \bibinfo{booktitle}{\emph{2022 IEEE Fourth
  International Conference on Advances in Electronics, Computers and
  Communications (ICAECC)}}. IEEE, \bibinfo{pages}{1--4}.
\newblock


\bibitem[Falcon et~al\mbox{.}(2022)]%
        {falcon2022image}
\bibfield{author}{\bibinfo{person}{Rafael Falcon}, \bibinfo{person}{Mauro
  Patti}, \bibinfo{person}{Stanislas Brochard-Garnier},
  \bibinfo{person}{G~Pacianotto Gouveia}, \bibinfo{person}{S~Torres Acevedo},
  \bibinfo{person}{Thelma Bergot}, \bibinfo{person}{Rick Alarcon},
  \bibinfo{person}{Corentin Bomstein}, \bibinfo{person}{Herv{\'e} Macudzinski},
  \bibinfo{person}{P Maitre}, {et~al\mbox{.}}} \bibinfo{year}{2022}\natexlab{}.
\newblock \showarticletitle{Image quality evaluation of video conferencing
  solutions with realistic laboratory scenes}.
\newblock \bibinfo{journal}{\emph{Electronic Imaging}} \bibinfo{volume}{34},
  \bibinfo{number}{9} (\bibinfo{year}{2022}), \bibinfo{pages}{318--1}.
\newblock


\bibitem[Feathers(2021)]%
        {feathers2021google}
\bibfield{author}{\bibinfo{person}{Todd Feathers}.}
  \bibinfo{year}{2021}\natexlab{}.
\newblock \showarticletitle{Google’s new dermatology app wasn’t designed
  for people with darker skin}.
\newblock \bibinfo{journal}{\emph{Retrieved August}}  \bibinfo{volume}{10}
  (\bibinfo{year}{2021}), \bibinfo{pages}{2022}.
\newblock


\bibitem[Fitzpatrick(1988a)]%
        {10.1001/archderm.1988.01670060015008}
\bibfield{author}{\bibinfo{person}{Thomas~B. Fitzpatrick}.}
  \bibinfo{year}{1988}\natexlab{a}.
\newblock \showarticletitle{{The Validity and Practicality of Sun-Reactive Skin
  Types I Through VI}}.
\newblock \bibinfo{journal}{\emph{Archives of Dermatology}}
  \bibinfo{volume}{124}, \bibinfo{number}{6} (\bibinfo{date}{06}
  \bibinfo{year}{1988}), \bibinfo{pages}{869--871}.
\newblock
\showISSN{0003-987X}
\urldef\tempurl%
\url{https://doi.org/10.1001/archderm.1988.01670060015008}
\showDOI{\tempurl}


\bibitem[Fitzpatrick(1988b)]%
        {fitzpatrick1988validity}
\bibfield{author}{\bibinfo{person}{Thomas~B Fitzpatrick}.}
  \bibinfo{year}{1988}\natexlab{b}.
\newblock \showarticletitle{The validity and practicality of sun-reactive skin
  types I through VI}.
\newblock \bibinfo{journal}{\emph{Archives of dermatology}}
  \bibinfo{volume}{124}, \bibinfo{number}{6} (\bibinfo{year}{1988}),
  \bibinfo{pages}{869--871}.
\newblock


\bibitem[Garcia et~al\mbox{.}(2018)]%
        {garcia2018pornographic}
\bibfield{author}{\bibinfo{person}{Manuel~B Garcia}, \bibinfo{person}{Teodoro~F
  Revano}, \bibinfo{person}{Beau Gray~M Habal}, \bibinfo{person}{Jennifer~O
  Contreras}, {and} \bibinfo{person}{John Benedic~R Enriquez}.}
  \bibinfo{year}{2018}\natexlab{}.
\newblock \showarticletitle{A pornographic image and video filtering
  application using optimized nudity recognition and detection algorithm}. In
  \bibinfo{booktitle}{\emph{2018 IEEE 10th International Conference on
  Humanoid, Nanotechnology, Information Technology, Communication and Control,
  Environment and Management (HNICEM)}}. IEEE, \bibinfo{pages}{1--5}.
\newblock


\bibitem[Gordon et~al\mbox{.}(2021)]%
        {gordon2021-disagreementdeconv}
\bibfield{author}{\bibinfo{person}{Mitchell~L. Gordon},
  \bibinfo{person}{Kaitlyn Zhou}, \bibinfo{person}{Kayur Patel},
  \bibinfo{person}{Tatsunori Hashimoto}, {and} \bibinfo{person}{Michael~S.
  Bernstein}.} \bibinfo{year}{2021}\natexlab{}.
\newblock \showarticletitle{The Disagreement Deconvolution: Bringing Machine
  Learning Performance Metrics In Line With Reality}. In
  \bibinfo{booktitle}{\emph{Proceedings of the 2021 CHI Conference on Human
  Factors in Computing Systems}} (Yokohama, Japan) \emph{(\bibinfo{series}{CHI
  '21})}. \bibinfo{publisher}{Association for Computing Machinery},
  \bibinfo{address}{New York, NY, USA}, Article \bibinfo{articleno}{388},
  \bibinfo{numpages}{14}~pages.
\newblock
\showISBNx{9781450380966}
\urldef\tempurl%
\url{https://doi.org/10.1145/3411764.3445423}
\showDOI{\tempurl}


\bibitem[Groh et~al\mbox{.}(2022)]%
        {groh2022towards}
\bibfield{author}{\bibinfo{person}{Matthew Groh}, \bibinfo{person}{Caleb
  Harris}, \bibinfo{person}{Roxana Daneshjou}, \bibinfo{person}{Omar Badri},
  {and} \bibinfo{person}{Arash Koochek}.} \bibinfo{year}{2022}\natexlab{}.
\newblock \showarticletitle{Towards transparency in dermatology image datasets
  with skin tone annotations by experts, crowds, and an algorithm}.
\newblock \bibinfo{journal}{\emph{arXiv preprint arXiv:2207.02942}}
  (\bibinfo{year}{2022}).
\newblock


\bibitem[Groh et~al\mbox{.}(2021)]%
        {groh2021evaluating}
\bibfield{author}{\bibinfo{person}{Matthew Groh}, \bibinfo{person}{Caleb
  Harris}, \bibinfo{person}{Luis Soenksen}, \bibinfo{person}{Felix Lau},
  \bibinfo{person}{Rachel Han}, \bibinfo{person}{Aerin Kim},
  \bibinfo{person}{Arash Koochek}, {and} \bibinfo{person}{Omar Badri}.}
  \bibinfo{year}{2021}\natexlab{}.
\newblock \showarticletitle{Evaluating deep neural networks trained on clinical
  images in dermatology with the fitzpatrick 17k dataset}. In
  \bibinfo{booktitle}{\emph{Proceedings of the IEEE/CVF Conference on Computer
  Vision and Pattern Recognition}}. \bibinfo{pages}{1820--1828}.
\newblock


\bibitem[G{\"u}{\c{c}}l{\"u}t{\"u}rk et~al\mbox{.}(2017)]%
        {guccluturk2017reconstructing}
\bibfield{author}{\bibinfo{person}{Ya{\u{g}}mur G{\"u}{\c{c}}l{\"u}t{\"u}rk},
  \bibinfo{person}{Umut G{\"u}{\c{c}}l{\"u}}, \bibinfo{person}{Katja Seeliger},
  \bibinfo{person}{Sander Bosch}, \bibinfo{person}{Rob van Lier}, {and}
  \bibinfo{person}{Marcel~A van Gerven}.} \bibinfo{year}{2017}\natexlab{}.
\newblock \showarticletitle{Reconstructing perceived faces from brain
  activations with deep adversarial neural decoding}.
\newblock \bibinfo{journal}{\emph{Advances in neural information processing
  systems}}  \bibinfo{volume}{30} (\bibinfo{year}{2017}).
\newblock


\bibitem[Haider and Sabahat(2022)]%
        {haider2022usability}
\bibfield{author}{\bibinfo{person}{Ammar Haider} {and} \bibinfo{person}{Nosheen
  Sabahat}.} \bibinfo{year}{2022}\natexlab{}.
\newblock \showarticletitle{A Usability and Accuracy Measurement of Smartphones
  Face Recognition}. In \bibinfo{booktitle}{\emph{2022 2nd International
  Conference on Artificial Intelligence (ICAI)}}. IEEE,
  \bibinfo{pages}{19--25}.
\newblock


\bibitem[Hazirbas et~al\mbox{.}(2021)]%
        {hazirbas2021towards}
\bibfield{author}{\bibinfo{person}{Caner Hazirbas}, \bibinfo{person}{Joanna
  Bitton}, \bibinfo{person}{Brian Dolhansky}, \bibinfo{person}{Jacqueline Pan},
  \bibinfo{person}{Albert Gordo}, {and} \bibinfo{person}{Cristian~Canton
  Ferrer}.} \bibinfo{year}{2021}\natexlab{}.
\newblock \showarticletitle{Towards measuring fairness in ai: the casual
  conversations dataset}.
\newblock \bibinfo{journal}{\emph{IEEE Transactions on Biometrics, Behavior,
  and Identity Science}} (\bibinfo{year}{2021}).
\newblock


\bibitem[Hoberman(2012)]%
        {hoberman2012black}
\bibfield{author}{\bibinfo{person}{John Hoberman}.}
  \bibinfo{year}{2012}\natexlab{}.
\newblock \bibinfo{booktitle}{\emph{Black and blue: The origins and
  consequences of medical racism}}.
\newblock \bibinfo{publisher}{Univ of California Press}.
\newblock


\bibitem[Howard et~al\mbox{.}(2021)]%
        {howard2021reliability}
\bibfield{author}{\bibinfo{person}{John~J Howard}, \bibinfo{person}{Yevgeniy~B
  Sirotin}, \bibinfo{person}{Jerry~L Tipton}, {and} \bibinfo{person}{Arun~R
  Vemury}.} \bibinfo{year}{2021}\natexlab{}.
\newblock \showarticletitle{Reliability and validity of image-based and
  self-reported skin phenotype metrics}.
\newblock \bibinfo{journal}{\emph{IEEE Transactions on Biometrics, Behavior,
  and Identity Science}} \bibinfo{volume}{3}, \bibinfo{number}{4}
  (\bibinfo{year}{2021}), \bibinfo{pages}{550--560}.
\newblock


\bibitem[Jo and Gebru(2020)]%
        {jo2020lessons}
\bibfield{author}{\bibinfo{person}{Eun~Seo Jo} {and} \bibinfo{person}{Timnit
  Gebru}.} \bibinfo{year}{2020}\natexlab{}.
\newblock \showarticletitle{Lessons from archives: Strategies for collecting
  sociocultural data in machine learning}. In
  \bibinfo{booktitle}{\emph{Proceedings of the 2020 conference on fairness,
  accountability, and transparency}}. \bibinfo{pages}{306--316}.
\newblock


\bibitem[Kazai et~al\mbox{.}(2012)]%
        {Kazai2012TheFO}
\bibfield{author}{\bibinfo{person}{Gabriella Kazai}, \bibinfo{person}{J.
  Kamps}, {and} \bibinfo{person}{Natasa Milic-Frayling}.}
  \bibinfo{year}{2012}\natexlab{}.
\newblock \showarticletitle{The face of quality in crowdsourcing relevance
  labels: demographics, personality and labeling accuracy}.
\newblock \bibinfo{journal}{\emph{Proceedings of the 21st ACM international
  conference on Information and knowledge management}} (\bibinfo{year}{2012}).
\newblock


\bibitem[Khan and Fu(2021)]%
        {khan2021one}
\bibfield{author}{\bibinfo{person}{Zaid Khan} {and} \bibinfo{person}{Yun Fu}.}
  \bibinfo{year}{2021}\natexlab{}.
\newblock \showarticletitle{One label, one billion faces: Usage and consistency
  of racial categories in computer vision}. In
  \bibinfo{booktitle}{\emph{Proceedings of the 2021 acm conference on fairness,
  accountability, and transparency}}. \bibinfo{pages}{587--597}.
\newblock


\bibitem[Kim et~al\mbox{.}(2022)]%
        {kim2022out}
\bibfield{author}{\bibinfo{person}{Hannah Kim}, \bibinfo{person}{Girmaw~Abebe
  Tadesse}, \bibinfo{person}{Celia Cintas}, \bibinfo{person}{Skyler Speakman},
  {and} \bibinfo{person}{Kush Varshney}.} \bibinfo{year}{2022}\natexlab{}.
\newblock \showarticletitle{Out-of-distribution detection in dermatology using
  input perturbation and subset scanning}. In \bibinfo{booktitle}{\emph{2022
  IEEE 19th International Symposium on Biomedical Imaging (ISBI)}}. IEEE,
  \bibinfo{pages}{1--4}.
\newblock


\bibitem[Kinyanjui et~al\mbox{.}(2019)]%
        {kinyanjui2019estimating}
\bibfield{author}{\bibinfo{person}{Newton~M Kinyanjui},
  \bibinfo{person}{Timothy Odonga}, \bibinfo{person}{Celia Cintas},
  \bibinfo{person}{Noel~CF Codella}, \bibinfo{person}{Rameswar Panda},
  \bibinfo{person}{Prasanna Sattigeri}, {and} \bibinfo{person}{Kush~R
  Varshney}.} \bibinfo{year}{2019}\natexlab{}.
\newblock \showarticletitle{Estimating skin tone and effects on classification
  performance in dermatology datasets}.
\newblock \bibinfo{journal}{\emph{arXiv preprint arXiv:1910.13268}}
  (\bibinfo{year}{2019}).
\newblock


\bibitem[Klare et~al\mbox{.}(2012)]%
        {klare2012face}
\bibfield{author}{\bibinfo{person}{Brendan~F Klare}, \bibinfo{person}{Mark~J
  Burge}, \bibinfo{person}{Joshua~C Klontz}, \bibinfo{person}{Richard W~Vorder
  Bruegge}, {and} \bibinfo{person}{Anil~K Jain}.}
  \bibinfo{year}{2012}\natexlab{}.
\newblock \showarticletitle{Face recognition performance: Role of demographic
  information}.
\newblock \bibinfo{journal}{\emph{IEEE Transactions on information forensics
  and security}} \bibinfo{volume}{7}, \bibinfo{number}{6}
  (\bibinfo{year}{2012}), \bibinfo{pages}{1789--1801}.
\newblock


\bibitem[Koshy et~al\mbox{.}(2021)]%
        {koshy2021complexion}
\bibfield{author}{\bibinfo{person}{Reeta Koshy}, \bibinfo{person}{Anisha
  Gharat}, \bibinfo{person}{Tejashri Wagh}, {and} \bibinfo{person}{Siddesh
  Sonawane}.} \bibinfo{year}{2021}\natexlab{}.
\newblock \showarticletitle{A Complexion based Outfit color recommender using
  Neural Networks}. In \bibinfo{booktitle}{\emph{2021 International Conference
  on Advances in Electrical, Computing, Communication and Sustainable
  Technologies (ICAECT)}}. IEEE, \bibinfo{pages}{1--7}.
\newblock


\bibitem[Krishnapriya et~al\mbox{.}(2022)]%
        {krishnapriya2022analysis}
\bibfield{author}{\bibinfo{person}{KS Krishnapriya}, \bibinfo{person}{Gabriella
  Pangelinan}, \bibinfo{person}{Michael~C King}, {and} \bibinfo{person}{Kevin~W
  Bowyer}.} \bibinfo{year}{2022}\natexlab{}.
\newblock \showarticletitle{Analysis of Manual and Automated Skin Tone
  Assignments}. In \bibinfo{booktitle}{\emph{Proceedings of the IEEE/CVF Winter
  Conference on Applications of Computer Vision}}. \bibinfo{pages}{429--438}.
\newblock


\bibitem[Laranjeira~da Silva et~al\mbox{.}(2022)]%
        {laranjeira2022seeing}
\bibfield{author}{\bibinfo{person}{Camila Laranjeira~da Silva},
  \bibinfo{person}{Jo{\~a}o Macedo}, \bibinfo{person}{Sandra Avila}, {and}
  \bibinfo{person}{Jefersson dos Santos}.} \bibinfo{year}{2022}\natexlab{}.
\newblock \showarticletitle{Seeing without looking: Analysis pipeline for child
  sexual abuse datasets}. In \bibinfo{booktitle}{\emph{2022 ACM Conference on
  Fairness, Accountability, and Transparency}}. \bibinfo{pages}{2189--2205}.
\newblock


\bibitem[Le et~al\mbox{.}(2020)]%
        {le2020anonfaces}
\bibfield{author}{\bibinfo{person}{Minh-Ha Le}, \bibinfo{person}{Md~Sakib~Nizam
  Khan}, \bibinfo{person}{Georgia Tsaloli}, \bibinfo{person}{Niklas Carlsson},
  {and} \bibinfo{person}{Sonja Buchegger}.} \bibinfo{year}{2020}\natexlab{}.
\newblock \showarticletitle{Anonfaces: Anonymizing faces adjusted to
  constraints on efficacy and security}. In
  \bibinfo{booktitle}{\emph{Proceedings of the 19th Workshop on Privacy in the
  Electronic Society}}. \bibinfo{pages}{87--100}.
\newblock


\bibitem[Lester et~al\mbox{.}(2020)]%
        {lester2020absence}
\bibfield{author}{\bibinfo{person}{JC Lester}, \bibinfo{person}{JL Jia},
  \bibinfo{person}{L Zhang}, \bibinfo{person}{GA Okoye}, {and}
  \bibinfo{person}{E Linos}.} \bibinfo{year}{2020}\natexlab{}.
\newblock \showarticletitle{Absence of images of skin of colour in publications
  of COVID-19 skin manifestations}.
\newblock \bibinfo{journal}{\emph{British Journal of Dermatology}}
  \bibinfo{volume}{183}, \bibinfo{number}{3} (\bibinfo{year}{2020}),
  \bibinfo{pages}{593--595}.
\newblock


\bibitem[Liu et~al\mbox{.}(2016)]%
        {liu2016effective}
\bibfield{author}{\bibinfo{person}{Angli Liu}, \bibinfo{person}{Stephen
  Soderland}, \bibinfo{person}{Jonathan Bragg}, \bibinfo{person}{Christopher~H
  Lin}, \bibinfo{person}{Xiao Ling}, {and} \bibinfo{person}{Daniel~S Weld}.}
  \bibinfo{year}{2016}\natexlab{}.
\newblock \showarticletitle{Effective crowd annotation for relation
  extraction}. In \bibinfo{booktitle}{\emph{Proceedings of the 2016 conference
  of the North American chapter of the association for computational
  linguistics: human language technologies}}. \bibinfo{pages}{897--906}.
\newblock


\bibitem[Liu et~al\mbox{.}(2021)]%
        {liu2021metaphys}
\bibfield{author}{\bibinfo{person}{Xin Liu}, \bibinfo{person}{Ziheng Jiang},
  \bibinfo{person}{Josh Fromm}, \bibinfo{person}{Xuhai Xu},
  \bibinfo{person}{Shwetak Patel}, {and} \bibinfo{person}{Daniel McDuff}.}
  \bibinfo{year}{2021}\natexlab{}.
\newblock \showarticletitle{MetaPhys: few-shot adaptation for non-contact
  physiological measurement}. In \bibinfo{booktitle}{\emph{Proceedings of the
  conference on health, inference, and learning}}. \bibinfo{pages}{154--163}.
\newblock


\bibitem[Liu et~al\mbox{.}(2015)]%
        {liu2015deep}
\bibfield{author}{\bibinfo{person}{Ziwei Liu}, \bibinfo{person}{Ping Luo},
  \bibinfo{person}{Xiaogang Wang}, {and} \bibinfo{person}{Xiaoou Tang}.}
  \bibinfo{year}{2015}\natexlab{}.
\newblock \showarticletitle{Deep learning face attributes in the wild}. In
  \bibinfo{booktitle}{\emph{Proceedings of the IEEE international conference on
  computer vision}}. \bibinfo{pages}{3730--3738}.
\newblock


\bibitem[Maduranga and Nandasena(2022)]%
        {maduranga2022mobile}
\bibfield{author}{\bibinfo{person}{MWP Maduranga} {and}
  \bibinfo{person}{Dilshan Nandasena}.} \bibinfo{year}{2022}\natexlab{}.
\newblock \showarticletitle{Mobile-based skin disease diagnosis system using
  convolutional neural networks (CNN)}.
\newblock \bibinfo{journal}{\emph{IJ Image Graphics Signal Process.}}
  \bibinfo{volume}{3} (\bibinfo{year}{2022}), \bibinfo{pages}{47--57}.
\newblock


\bibitem[Maze et~al\mbox{.}(2018)]%
        {maze2018iarpa}
\bibfield{author}{\bibinfo{person}{Brianna Maze}, \bibinfo{person}{Jocelyn
  Adams}, \bibinfo{person}{James~A Duncan}, \bibinfo{person}{Nathan Kalka},
  \bibinfo{person}{Tim Miller}, \bibinfo{person}{Charles Otto},
  \bibinfo{person}{Anil~K Jain}, \bibinfo{person}{W~Tyler Niggel},
  \bibinfo{person}{Janet Anderson}, \bibinfo{person}{Jordan Cheney},
  {et~al\mbox{.}}} \bibinfo{year}{2018}\natexlab{}.
\newblock \showarticletitle{Iarpa janus benchmark-c: Face dataset and
  protocol}. In \bibinfo{booktitle}{\emph{2018 international conference on
  biometrics (ICB)}}. IEEE, \bibinfo{pages}{158--165}.
\newblock


\bibitem[McDuff et~al\mbox{.}(2021)]%
        {mcduff2021synthetic}
\bibfield{author}{\bibinfo{person}{Daniel McDuff}, \bibinfo{person}{Xin Liu},
  \bibinfo{person}{Javier Hernandez}, \bibinfo{person}{Erroll Wood}, {and}
  \bibinfo{person}{Tadas Baltrusaitis}.} \bibinfo{year}{2021}\natexlab{}.
\newblock \showarticletitle{Synthetic Data for Multi-Parameter Camera-Based
  Physiological Sensing}. In \bibinfo{booktitle}{\emph{2021 43rd Annual
  International Conference of the IEEE Engineering in Medicine \& Biology
  Society (EMBC)}}. IEEE, \bibinfo{pages}{3742--3748}.
\newblock


\bibitem[McDuff et~al\mbox{.}(2019)]%
        {mcduff2019characterizing}
\bibfield{author}{\bibinfo{person}{Daniel McDuff}, \bibinfo{person}{Shuang Ma},
  \bibinfo{person}{Yale Song}, {and} \bibinfo{person}{Ashish Kapoor}.}
  \bibinfo{year}{2019}\natexlab{}.
\newblock \showarticletitle{Characterizing bias in classifiers using generative
  models}.
\newblock \bibinfo{journal}{\emph{Advances in neural information processing
  systems}}  \bibinfo{volume}{32} (\bibinfo{year}{2019}).
\newblock


\bibitem[Mehrotra and Celis(2021)]%
        {mehrotra2021mitigating}
\bibfield{author}{\bibinfo{person}{Anay Mehrotra} {and}
  \bibinfo{person}{L~Elisa Celis}.} \bibinfo{year}{2021}\natexlab{}.
\newblock \showarticletitle{Mitigating bias in set selection with noisy
  protected attributes}. In \bibinfo{booktitle}{\emph{Proceedings of the 2021
  ACM Conference on Fairness, Accountability, and Transparency}}.
  \bibinfo{pages}{237--248}.
\newblock


\bibitem[Menezes et~al\mbox{.}(2021)]%
        {menezes2021bias}
\bibfield{author}{\bibinfo{person}{Hanna~F Menezes}, \bibinfo{person}{Arthur~SC
  Ferreira}, \bibinfo{person}{Eanes~T Pereira}, {and} \bibinfo{person}{Herman~M
  Gomes}.} \bibinfo{year}{2021}\natexlab{}.
\newblock \showarticletitle{Bias and Fairness in Face Detection}. In
  \bibinfo{booktitle}{\emph{2021 34th SIBGRAPI Conference on Graphics, Patterns
  and Images (SIBGRAPI)}}. IEEE, \bibinfo{pages}{247--254}.
\newblock


\bibitem[Menon(2023)]%
        {menon2023leveraging}
\bibfield{author}{\bibinfo{person}{Arjun Menon}.}
  \bibinfo{year}{2023}\natexlab{}.
\newblock \showarticletitle{Leveraging Facial Recognition Technology in
  Criminal Identification}.
\newblock \bibinfo{journal}{\emph{Interdisciplinary Innovations and
  Developments towards Smart and Sustainable Industries https://doi.
  org/10.13052/rp-978-87-7022-828-2}} (\bibinfo{year}{2023}).
\newblock


\bibitem[Merler et~al\mbox{.}(2019)]%
        {merler2019diversity}
\bibfield{author}{\bibinfo{person}{Michele Merler}, \bibinfo{person}{Nalini
  Ratha}, \bibinfo{person}{Rogerio~S Feris}, {and} \bibinfo{person}{John~R
  Smith}.} \bibinfo{year}{2019}\natexlab{}.
\newblock \showarticletitle{Diversity in faces}.
\newblock \bibinfo{journal}{\emph{arXiv preprint arXiv:1901.10436}}
  (\bibinfo{year}{2019}).
\newblock


\bibitem[Mishra et~al\mbox{.}(2021)]%
        {mishra2021dual}
\bibfield{author}{\bibinfo{person}{Shiksha Mishra}, \bibinfo{person}{Puspita
  Majumdar}, \bibinfo{person}{Muskan Dosi}, \bibinfo{person}{Mayank Vatsa},
  {and} \bibinfo{person}{Richa Singh}.} \bibinfo{year}{2021}\natexlab{}.
\newblock \showarticletitle{Dual sensor indian masked face dataset}. In
  \bibinfo{booktitle}{\emph{2021 16th IEEE International Conference on
  Automatic Face and Gesture Recognition (FG 2021)}}. IEEE,
  \bibinfo{pages}{1--8}.
\newblock


\bibitem[Molina et~al\mbox{.}(2020)]%
        {molina2020reduction}
\bibfield{author}{\bibinfo{person}{David~A Molina}, \bibinfo{person}{Leonardo
  Causa}, {and} \bibinfo{person}{Juan Tapia}.} \bibinfo{year}{2020}\natexlab{}.
\newblock \showarticletitle{Reduction of bias for gender and ethnicity from
  face images using automated skin tone classification}. In
  \bibinfo{booktitle}{\emph{2020 International Conference of the Biometrics
  Special Interest Group (BIOSIG)}}. IEEE, \bibinfo{pages}{1--5}.
\newblock


\bibitem[Monk(2019)]%
        {monk2019color}
\bibfield{author}{\bibinfo{person}{Ellis~P Monk}.}
  \bibinfo{year}{2019}\natexlab{}.
\newblock \showarticletitle{The color of punishment: African Americans, skin
  tone, and the criminal justice system}.
\newblock \bibinfo{journal}{\emph{Ethnic and Racial Studies}}
  \bibinfo{volume}{42}, \bibinfo{number}{10} (\bibinfo{year}{2019}),
  \bibinfo{pages}{1593--1612}.
\newblock


\bibitem[Monk~Jr(2021)]%
        {monk2021unceasing}
\bibfield{author}{\bibinfo{person}{Ellis~P Monk~Jr}.}
  \bibinfo{year}{2021}\natexlab{}.
\newblock \showarticletitle{The unceasing significance of colorism: Skin tone
  stratification in the United States}.
\newblock \bibinfo{journal}{\emph{Daedalus}} \bibinfo{volume}{150},
  \bibinfo{number}{2} (\bibinfo{year}{2021}), \bibinfo{pages}{76--90}.
\newblock


\bibitem[Monk~Jr et~al\mbox{.}(2021)]%
        {monk2021skin}
\bibfield{author}{\bibinfo{person}{Ellis~P Monk~Jr}, \bibinfo{person}{Jerry
  Kaufman}, {and} \bibinfo{person}{Yadira Montoya}.}
  \bibinfo{year}{2021}\natexlab{}.
\newblock \showarticletitle{Skin tone and perceived discrimination: Health and
  aging beyond the binary in NSHAP 2015}.
\newblock \bibinfo{journal}{\emph{The Journals of Gerontology: Series B}}
  \bibinfo{volume}{76}, \bibinfo{number}{Supplement\_3} (\bibinfo{year}{2021}),
  \bibinfo{pages}{S313--S321}.
\newblock


\bibitem[Muthukumar(2019)]%
        {muthukumar2019color}
\bibfield{author}{\bibinfo{person}{Vidya Muthukumar}.}
  \bibinfo{year}{2019}\natexlab{}.
\newblock \showarticletitle{Color-theoretic experiments to understand unequal
  gender classification accuracy from face images}. In
  \bibinfo{booktitle}{\emph{Proceedings of the IEEE/CVF Conference on Computer
  Vision and Pattern Recognition Workshops}}. \bibinfo{pages}{0--0}.
\newblock


\bibitem[O'neil(2017)]%
        {o2017weapons}
\bibfield{author}{\bibinfo{person}{Cathy O'neil}.}
  \bibinfo{year}{2017}\natexlab{}.
\newblock \bibinfo{booktitle}{\emph{Weapons of math destruction: How big data
  increases inequality and threatens democracy}}.
\newblock \bibinfo{publisher}{Crown}.
\newblock


\bibitem[Park et~al\mbox{.}(2021)]%
        {park2021reliable}
\bibfield{author}{\bibinfo{person}{Chunjong Park}, \bibinfo{person}{Anas
  Awadalla}, \bibinfo{person}{Tadayoshi Kohno}, {and} \bibinfo{person}{Shwetak
  Patel}.} \bibinfo{year}{2021}\natexlab{}.
\newblock \showarticletitle{Reliable and trustworthy machine learning for
  health using dataset shift detection}.
\newblock \bibinfo{journal}{\emph{Advances in Neural Information Processing
  Systems}}  \bibinfo{volume}{34} (\bibinfo{year}{2021}),
  \bibinfo{pages}{3043--3056}.
\newblock


\bibitem[Park et~al\mbox{.}(2018)]%
        {park2018automatic}
\bibfield{author}{\bibinfo{person}{Jisoo Park}, \bibinfo{person}{Hyungjoon
  Kim}, \bibinfo{person}{Seonmi Ji}, {and} \bibinfo{person}{Eenjun Hwang}.}
  \bibinfo{year}{2018}\natexlab{}.
\newblock \showarticletitle{An automatic virtual makeup scheme based on
  personal color analysis}. In \bibinfo{booktitle}{\emph{Proceedings of the
  12th International Conference on Ubiquitous Information Management and
  Communication}}. \bibinfo{pages}{1--7}.
\newblock


\bibitem[Perera et~al\mbox{.}(2021)]%
        {perera2021virtual}
\bibfield{author}{\bibinfo{person}{PRH Perera}, \bibinfo{person}{ESS Soysa},
  \bibinfo{person}{HRS De~Silva}, \bibinfo{person}{ARP Tavarayan},
  \bibinfo{person}{MP Gamage}, {and} \bibinfo{person}{KMLP Weerasinghe}.}
  \bibinfo{year}{2021}\natexlab{}.
\newblock \showarticletitle{Virtual Makeover and Makeup Recommendation Based on
  Personal Trait Analysis}. In \bibinfo{booktitle}{\emph{2021 3rd International
  Conference on Advancements in Computing (ICAC)}}. IEEE,
  \bibinfo{pages}{288--293}.
\newblock


\bibitem[Research(2022)]%
        {monkscale}
\bibfield{author}{\bibinfo{person}{Google Research}.}
  \bibinfo{year}{2022}\natexlab{}.
\newblock \bibinfo{title}{Developing the Monk Skin Tone Scale}.
\newblock
\newblock
\newblock
\shownote{\url{https://skintone.google/the-scale}}.


\bibitem[Ross et~al\mbox{.}(2010)]%
        {Ross2010WhoAT}
\bibfield{author}{\bibinfo{person}{Joel Ross}, \bibinfo{person}{Lilly~C.
  Irani}, \bibinfo{person}{M.~Six Silberman}, \bibinfo{person}{Andrew
  Zaldivar}, {and} \bibinfo{person}{Bill Tomlinson}.}
  \bibinfo{year}{2010}\natexlab{}.
\newblock \showarticletitle{Who are the crowdworkers?: shifting demographics in
  mechanical turk}.
\newblock \bibinfo{journal}{\emph{CHI '10 Extended Abstracts on Human Factors
  in Computing Systems}} (\bibinfo{year}{2010}).
\newblock


\bibitem[Ryan-Mosley(2021)]%
        {ryan2021new}
\bibfield{author}{\bibinfo{person}{Tate Ryan-Mosley}.}
  \bibinfo{year}{2021}\natexlab{}.
\newblock \bibinfo{title}{The new lawsuit that shows facial recognition is
  officially a civil rights issue}.
\newblock
\newblock
\newblock
\shownote{\url{https://www.technologyreview.com/2021/04/14/1022676/robert-williams-facial-recognition-lawsuit-aclu-detroit-police/}}.


\bibitem[Scheuerman et~al\mbox{.}(2020)]%
        {scheuerman2020we}
\bibfield{author}{\bibinfo{person}{Morgan~Klaus Scheuerman},
  \bibinfo{person}{Kandrea Wade}, \bibinfo{person}{Caitlin Lustig}, {and}
  \bibinfo{person}{Jed~R Brubaker}.} \bibinfo{year}{2020}\natexlab{}.
\newblock \showarticletitle{How we've taught algorithms to see identity:
  Constructing race and gender in image databases for facial analysis}.
\newblock \bibinfo{journal}{\emph{Proceedings of the ACM on Human-computer
  Interaction}} \bibinfo{volume}{4}, \bibinfo{number}{CSCW1}
  (\bibinfo{year}{2020}), \bibinfo{pages}{1--35}.
\newblock


\bibitem[Shah et~al\mbox{.}(2022)]%
        {shah2022deep}
\bibfield{author}{\bibinfo{person}{Zaineb Shah}, \bibinfo{person}{Syed Ayaz~Ali
  Shah}, \bibinfo{person}{Aamir Shahzad}, \bibinfo{person}{Ahmad Fayyaz},
  \bibinfo{person}{Shoaib Khaliq}, \bibinfo{person}{Ali Zahir}, {and}
  \bibinfo{person}{Goh~Chuan Meng}.} \bibinfo{year}{2022}\natexlab{}.
\newblock \showarticletitle{Deep Learning-Based Forearm Subcutaneous Veins
  Segmentation}.
\newblock \bibinfo{journal}{\emph{IEEE Access}}  \bibinfo{volume}{10}
  (\bibinfo{year}{2022}), \bibinfo{pages}{42814--42820}.
\newblock


\bibitem[Sixta et~al\mbox{.}(2020)]%
        {sixta2020fairface}
\bibfield{author}{\bibinfo{person}{Tom{\'a}{\v{s}} Sixta},
  \bibinfo{person}{Julio~CS Jacques~Junior}, \bibinfo{person}{Pau
  Buch-Cardona}, \bibinfo{person}{Eduard Vazquez}, {and}
  \bibinfo{person}{Sergio Escalera}.} \bibinfo{year}{2020}\natexlab{}.
\newblock \showarticletitle{Fairface challenge at eccv 2020: Analyzing bias in
  face recognition}. In \bibinfo{booktitle}{\emph{Computer Vision--ECCV 2020
  Workshops: Glasgow, UK, August 23--28, 2020, Proceedings, Part VI 16}}.
  Springer, \bibinfo{pages}{463--481}.
\newblock


\bibitem[Spetl{\'\i}k et~al\mbox{.}(2018)]%
        {spetlik2018non}
\bibfield{author}{\bibinfo{person}{Radim Spetl{\'\i}k}, \bibinfo{person}{Jan
  Cech}, {and} \bibinfo{person}{Jiri Matas}.} \bibinfo{year}{2018}\natexlab{}.
\newblock \showarticletitle{Non-contact reflectance photoplethysmography:
  Progress, limitations, and myths}. In \bibinfo{booktitle}{\emph{2018 13th
  IEEE International Conference on Automatic Face \& Gesture Recognition (FG
  2018)}}. IEEE, \bibinfo{pages}{702--709}.
\newblock


\bibitem[Swauger(2020)]%
        {swauger2020our}
\bibfield{author}{\bibinfo{person}{Shea Swauger}.}
  \bibinfo{year}{2020}\natexlab{}.
\newblock \showarticletitle{Our bodies encoded: Algorithmic test proctoring in
  higher education}.
\newblock \bibinfo{journal}{\emph{Critical digital pedagogy}}
  (\bibinfo{year}{2020}).
\newblock


\bibitem[Tariq et~al\mbox{.}(2018)]%
        {tariq2018designing}
\bibfield{author}{\bibinfo{person}{Muhammad~Uzair Tariq},
  \bibinfo{person}{Arup~Kumar Ghosh}, \bibinfo{person}{Karla Badillo-Urquiola},
  \bibinfo{person}{Abhiditya Jha}, \bibinfo{person}{Sanjeev Koppal}, {and}
  \bibinfo{person}{Pamela~J Wisniewski}.} \bibinfo{year}{2018}\natexlab{}.
\newblock \bibinfo{booktitle}{\emph{Designing light filters to detect skin
  using a low-powered sensor}}.
\newblock \bibinfo{publisher}{IEEE}.
\newblock


\bibitem[Toyoda et~al\mbox{.}(2021)]%
        {toyoda2021predicting}
\bibfield{author}{\bibinfo{person}{Yuushi Toyoda}, \bibinfo{person}{Gale
  Lucas}, {and} \bibinfo{person}{Jonathan Gratch}.}
  \bibinfo{year}{2021}\natexlab{}.
\newblock \showarticletitle{Predicting Worker Accuracy from Nonverbal
  Behaviour: Benefits and Potential for Algorithmic Bias}. In
  \bibinfo{booktitle}{\emph{Companion Publication of the 2021 International
  Conference on Multimodal Interaction}}. \bibinfo{pages}{25--30}.
\newblock


\bibitem[Vitale(2021)]%
        {vitale2021end}
\bibfield{author}{\bibinfo{person}{Alex~S Vitale}.}
  \bibinfo{year}{2021}\natexlab{}.
\newblock \bibinfo{booktitle}{\emph{The end of policing}}.
\newblock \bibinfo{publisher}{Verso Books}.
\newblock


\bibitem[Wang et~al\mbox{.}(2016)]%
        {wang2016algorithmic}
\bibfield{author}{\bibinfo{person}{Wenjin Wang}, \bibinfo{person}{Albertus~C
  Den~Brinker}, \bibinfo{person}{Sander Stuijk}, {and} \bibinfo{person}{Gerard
  De~Haan}.} \bibinfo{year}{2016}\natexlab{}.
\newblock \showarticletitle{Algorithmic principles of remote PPG}.
\newblock \bibinfo{journal}{\emph{IEEE Transactions on Biomedical Engineering}}
  \bibinfo{volume}{64}, \bibinfo{number}{7} (\bibinfo{year}{2016}),
  \bibinfo{pages}{1479--1491}.
\newblock


\bibitem[Ware et~al\mbox{.}(2020)]%
        {ware2020racial}
\bibfield{author}{\bibinfo{person}{Olivia~R Ware}, \bibinfo{person}{Jessica~E
  Dawson}, \bibinfo{person}{Michi~M Shinohara}, {and} \bibinfo{person}{Susan~C
  Taylor}.} \bibinfo{year}{2020}\natexlab{}.
\newblock \showarticletitle{Racial limitations of Fitzpatrick skin type}.
\newblock \bibinfo{journal}{\emph{Cutis}} \bibinfo{volume}{105},
  \bibinfo{number}{2} (\bibinfo{year}{2020}), \bibinfo{pages}{77--80}.
\newblock


\bibitem[Whitelam et~al\mbox{.}(2017)]%
        {whitelam2017iarpa}
\bibfield{author}{\bibinfo{person}{Cameron Whitelam}, \bibinfo{person}{Emma
  Taborsky}, \bibinfo{person}{Austin Blanton}, \bibinfo{person}{Brianna Maze},
  \bibinfo{person}{Jocelyn Adams}, \bibinfo{person}{Tim Miller},
  \bibinfo{person}{Nathan Kalka}, \bibinfo{person}{Anil~K Jain},
  \bibinfo{person}{James~A Duncan}, \bibinfo{person}{Kristen Allen},
  {et~al\mbox{.}}} \bibinfo{year}{2017}\natexlab{}.
\newblock \showarticletitle{Iarpa janus benchmark-b face dataset}. In
  \bibinfo{booktitle}{\emph{proceedings of the IEEE conference on computer
  vision and pattern recognition workshops}}. \bibinfo{pages}{90--98}.
\newblock


\bibitem[Wilson et~al\mbox{.}(2019)]%
        {wilson2019predictive}
\bibfield{author}{\bibinfo{person}{Benjamin Wilson}, \bibinfo{person}{Judy
  Hoffman}, {and} \bibinfo{person}{Jamie Morgenstern}.}
  \bibinfo{year}{2019}\natexlab{}.
\newblock \showarticletitle{Predictive inequity in object detection}.
\newblock \bibinfo{journal}{\emph{arXiv preprint arXiv:1902.11097}}
  (\bibinfo{year}{2019}).
\newblock


\bibitem[Yucer et~al\mbox{.}(2022)]%
        {yucer2022measuring}
\bibfield{author}{\bibinfo{person}{Seyma Yucer}, \bibinfo{person}{Furkan
  Tektas}, \bibinfo{person}{Noura Al~Moubayed}, {and} \bibinfo{person}{Toby~P
  Breckon}.} \bibinfo{year}{2022}\natexlab{}.
\newblock \showarticletitle{Measuring hidden bias within face recognition via
  racial phenotypes}. In \bibinfo{booktitle}{\emph{Proceedings of the IEEE/CVF
  Winter Conference on Applications of Computer Vision}}.
  \bibinfo{pages}{995--1004}.
\newblock


\bibitem[Zhang et~al\mbox{.}(2020)]%
        {zhang2020open}
\bibfield{author}{\bibinfo{person}{Longhao Zhang}, \bibinfo{person}{Xipeng
  Pan}, \bibinfo{person}{Huihua Yang}, {and} \bibinfo{person}{Lingqiao Li}.}
  \bibinfo{year}{2020}\natexlab{}.
\newblock \showarticletitle{On Open-Set, High-Fidelity and Identity-Specific
  Face Transformation}.
\newblock \bibinfo{journal}{\emph{IEEE Access}}  \bibinfo{volume}{8}
  (\bibinfo{year}{2020}), \bibinfo{pages}{224643--224653}.
\newblock


\bibitem[Zhang et~al\mbox{.}(2019)]%
        {zhang2019analysis}
\bibfield{author}{\bibinfo{person}{Xiaobiao Zhang}, \bibinfo{person}{Xiaoyi
  Feng}, {and} \bibinfo{person}{Zhaoqiang Xia}.}
  \bibinfo{year}{2019}\natexlab{}.
\newblock \showarticletitle{Analysis of Factors on BVP Signal Extraction Based
  on Imaging Principle}. In \bibinfo{booktitle}{\emph{Proceedings of the 2019
  3rd International Conference on Biometric Engineering and Applications}}.
  \bibinfo{pages}{48--55}.
\newblock


\bibitem[Zheng et~al\mbox{.}(2022)]%
        {zheng2022automatic}
\bibfield{author}{\bibinfo{person}{Qian Zheng}, \bibinfo{person}{Ankur Purwar},
  \bibinfo{person}{Heng Zhao}, \bibinfo{person}{Guang~Liang Lim},
  \bibinfo{person}{Ling Li}, \bibinfo{person}{Debasish Behera},
  \bibinfo{person}{Qian Wang}, \bibinfo{person}{Min Tan},
  \bibinfo{person}{Rizhao Cai}, \bibinfo{person}{Jennifer Werner},
  {et~al\mbox{.}}} \bibinfo{year}{2022}\natexlab{}.
\newblock \showarticletitle{Automatic facial skin feature detection for
  everyone}.
\newblock \bibinfo{journal}{\emph{arXiv preprint arXiv:2203.16056}}
  (\bibinfo{year}{2022}).
\newblock


\end{thebibliography}
